\documentclass{article} 
\usepackage{iclr2027_conference,times}

\usepackage{amsmath,amsfonts,bm}

\def\eqref#1{equation~\ref{#1}}

\def\1{\bm{1}}

\DeclareMathAlphabet{\mathsfit}{\encodingdefault}{\sfdefault}{m}{sl}
\SetMathAlphabet{\mathsfit}{bold}{\encodingdefault}{\sfdefault}{bx}{n}

\usepackage{hyperref}
\usepackage{url}
\usepackage{graphicx}

\usepackage{booktabs,amsmath,amssymb,xcolor}
\usepackage{fvextra,array,placeins,wrapfig,float}
\usepackage{tcolorbox,longtable,multirow}
\usepackage{enumitem}
\setlist[itemize]{nosep,rightmargin=2em}
\tcbuselibrary{breakable,skins}
\hypersetup{hidelinks,pdfauthor={},pdfsubject={}}
\newcommand{\ours}{\textsc{DataSense-Bench}}
\title{\ours: The First Step Toward \\an AI Scientist}

\author{\small Yudi Zhang$^{1}$\thanks{Work done during Yudi's internship at the Max Planck Institute for Intelligent Systems.}, Mingyu Cao$^{2}$, Lu Yin$^{2}$, Mykola Pechenizkiy$^{1}$, Shiwei Liu$^{3,4,5}$\\[3pt]
\normalfont\small $^{1}$Eindhoven University of Technology\quad
\normalfont\small $^{2}$University of Surrey\quad $^{3}$ELLIS Institute T\"ubingen\\
\normalfont\small $^{4}$Max Planck Institute for Intelligent Systems\quad
\normalfont\small $^{5}$T\"ubingen AI Center}

\iclrfinalcopy
\usepackage{etoolbox}
\usepackage{caption}
\makeatletter
\patchcmd{\@maketitle}{Published as a conference paper at ICLR 2027}{Preprint}{}{\PackageError{preprint}{Header replacement failed}{}}
\makeatother
\begin{document}

\maketitle
\vspace{-18pt}

\begin{abstract}
As claims about recursive self-improvement (RSI) and artificial general intelligence (AGI) proliferate, we ask a simple question: do frontier AI models have a sense of data, i.e., can they reliably select the right data for training? We introduce DataSense-Bench\footnote{\href{https://github.com/DataSense-Bench/DataSense-Bench}{\textcolor{blue}{Code}} and \href{https://datasense-bench.github.io/}{\textcolor{blue}{Project page}}.} to study this capability through the fundamental problem of data selection and performance forecasting in machine learning. We ask AI agents to select and rank candidate training subsets that can be used to fine-tune a small LLM model. Agents are allowed to inspect the data, write and execute analysis code, and run model forward passes, but can not train the model or access the actual evaluation tasks. We then fine-tune the base model on each selected subset and evaluate its post-training performance under a standardized protocol. We instantiate the benchmark in terminal problem solving and tool use, selecting trajectories from OpenThoughts-Agent and EnvScaler and evaluating on TBLite and BFCL, respectively. We then evaluate the agents along two complementary dimensions: the \textit{\textbf{post-training performance of the top-ranked subset}}, reflecting the ability to identify high-value training data, and \textit{\textbf{ranking accuracy}}, reflecting the ability to predict the relative performance of the selected subsets.
In our experiments, selection gains over random selection are limited; agents do not reliably rank their selected groups, and ranking ability does not hold consistently across tasks: Astra identifies the best group in all three tool-use runs but in only one of three terminal runs. Analysis of execution traces on both tasks shows that agents often use similar data signals while interpreting their training value differently.

\end{abstract}
\vspace{-8pt}
\begin{figure}[H]
\centering
\includegraphics[width=0.94\linewidth]{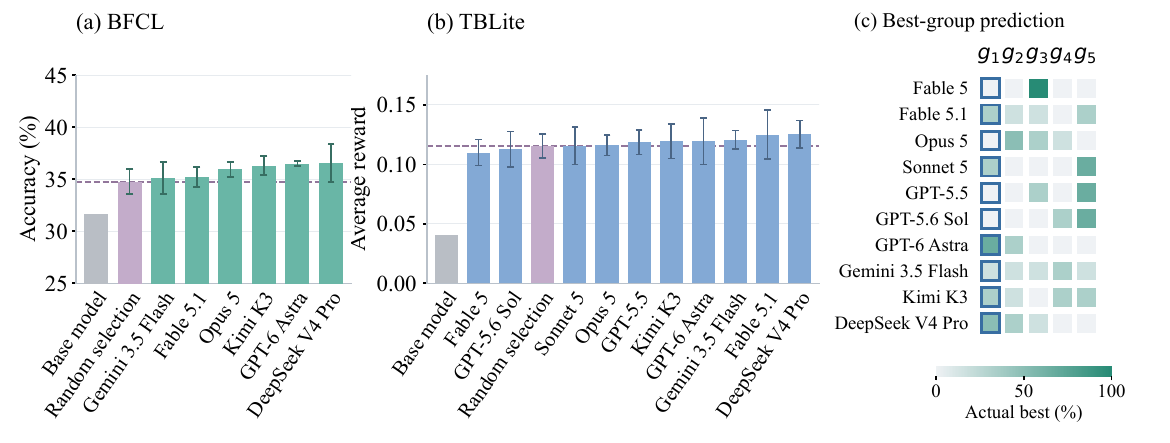}
\vspace{-5pt}
\begingroup\small
\caption{\textbf{Can CLI agents select useful training data and predict its value?}
\textbf{(a,b)} Qwen3-4B performance after training on the predicted top-ranked group selected by each agent.
Panel (a) reports BFCL accuracy; panel (b) reports TBLite reward.
Dashed lines denote random selection. Scores are averaged across evaluation seeds
within checkpoints, then across three agent runs; error bars show standard
deviation across agent runs (across evaluation seeds for random selection).
\textbf{(c)} Frequency of each predicted rank yielding the best group, pooling evaluated runs
from both tasks with equal weight per run: six runs for agents evaluated on both
tasks and three for TBLite-only agents. Tied best groups each count as correct;
blue outlines mark $g_1$.}
\label{fig:joint-results}
\endgroup
\end{figure}

\section{Introduction}

Large language models (LLMs) are increasingly used for tasks such as question answering~\citep{brown2020fewshot}, code generation~\citep{chen2021code}, and data analysis~\citep{discoverybench}. Recent advances in reasoning and tool use have made frontier models capable of tackling complex scientific problems, supporting AI scientists who generate research ideas, conduct experiments, analyze results, and write complete research manuscripts~\citep{aiscientist}. Benchmarks increasingly assess scientific programming, discovery, and machine learning experimentation~\citep{scienceagentbench,discoverybench,mlagentbench,posttrainbench}. These settings test several abilities together. They leave a more focused question open: can an AI system anticipate the value of an experimental choice before observing its outcome?

Training-data selection provides a concrete setting for this question. Before committing to training, researchers must choose which examples will help a particular model learn~\citep{lima,less}. This is a consequential and difficult decision: data selection determines the supervision a model receives, yet observable quality does not directly establish training value. For example, low loss alone need not imply high training value: an example may already be well learned~\citep{rholoss}. Likewise, a successful trajectory does not by itself establish how much a particular model will learn from it. Reliable judgment requires connecting properties of the data to what a particular model will learn under a fixed training procedure~\citep{ifd,less}.

\textbf{Our Benchmark.} Therefore, we introduce \ours{} to evaluate this judgment capability of the frontier models. A CLI agent receives a candidate training data pool, inference-only access to a base model, and a fixed fine-tuning protocol. It selects disjoint, equal-sized subsets and ranks them by expected value for the post-training performance. The agent may inspect data, execute analysis code, and run forward passes, but cannot train models or access evaluation tasks. Once the submission is fixed, the benchmark fine-tunes a fresh copy of the base model on each subset and measures the resulting performance. This separation makes the submitted ranking a prediction that can be checked against subsequent experimental evidence. As a consequence, the benchmark asks about both \emph{\textbf{Selection quality}} and \emph{\textbf{Ranking accuracy}}, which measure whether the first group in the LLM's predicted ordering produces a useful model and whether the predicted ordering of the subsets agrees with their observed performance, separately. A strong top choice need not imply an accurate ordering; conversely, correctly ordering weak subsets need not produce a useful recommendation. Both are needed to understand whether an agent can turn data analysis into reliable decisions.

\textbf{Our Findings.} We instantiate this protocol in terminal problem solving and multi-turn tool use, using OpenThoughts-Agent and EnvScaler trajectories for training and TBLite and BFCL for downstream evaluation, respectively. Our experiments reveal two limitations: gains over random selection are modest, and agents do not reliably rank their selected groups. Ranking ability also does not hold consistently across tasks. \textbf{(1) The frontier agents produce plausible, data-specific recipes but do not reliably predict post-training utility.} 
On TBLite, two LLMs select first groups that underperform random selection,
with differences across LLMs ranging from -0.55 
to +0.99 points, 
as shown in Table~\ref{tab:main-results}.
On BFCL, all six agents’ first groups exceed random selection by 0.35–1.78 \% points. Astra identifies the best group in all three BFCL runs but in only one of three TBLite runs. None of the six agents evaluated on both tasks identifies the best group in all three runs on each task. \textbf{(2) Selection recipes vary substantially across three independent runs of the same agent.} 
Astra’s TBLite runs change their quality-assessment procedures, while Kimi’s BFCL runs move from loss-outlier rejection to intermediate-loss and then lower-loss preferences. The selection rule itself changes across runs. \textbf{(3) Agents draw different conclusions from similar data signals.} For example, agents use base-model loss in substantially different ways across runs and tasks. Section 4.3 examines the concrete recipes in detail.
\section{Related Work}
\label{sec:related}
We review benchmarks for CLI agents, training-data selection, and performance forecasting.

\textbf{Benchmarks for CLI Agents.}
Command-line interface (CLI) agents use shell commands and scripts to
complete tasks in OS environments. Terminal-Bench evaluates
realistic terminal tasks with executable verification, while SWE-bench
focuses on resolving real GitHub issues through repository
modifications~\citep{terminalbench,swebench}. Research-oriented benchmarks
address scientific programming and
reproducibility~\citep{scienceagentbench,corebench}, as well as machine
learning experimentation and research
engineering~\citep{mlagentbench,mlebench,rebench}. Most closely related,
PostTrainBench evaluates agents' choices of data, methods, and
hyperparameters for LLM post-training under a compute
budget~\citep{posttrainbench}. These benchmarks measure end-to-end outcomes
that depend on both decision quality and execution. \ours{} focuses on
data-selection judgment: agents inspect candidate data, construct
selection procedures, and submit ranked subsets for a fixed base model
and training protocol before receiving any training feedback.

\textbf{Training Data Selection.}
Existing selection methods differ in the signals they use to identify
useful training data. Curation and scoring approaches emphasize the value
of small, high-quality instruction sets~\citep{lima} and use model-based
assessments of quality or instruction-following difficulty to select
examples~\citep{ICLR2024_9543942c,ifd}. Other approaches seek to use importance resampling~\citep{dsir} or improve coverage through deduplication and
embedding-based data selection~\citep{d4}. RHO-LOSS uses a holdout-trained model to
prioritize examples expected to reduce generalization loss~\citep{rholoss}.  More recently, FEEDER assesses the sufficiency and necessity
of demonstrations for in-context learning, with an extension to
fine-tuning~\citep{feeder}.  Together, these methods motivate the
quality, loss, and coverage signals examined in our procedure analysis.

\textbf{Performance Forecasting.} A complementary line estimates how training examples affect model
behavior through derivative-based influence
approximations~\citep{influence,tracin,datainf}, marginal contributions to
predictive performance~\citep{datashapley}, or gradient similarity for
targeted instruction tuning~\citep{less}. At the subset or mixture level,
datamodels predict model outputs from training-subset
membership~\citep{datamodels}, while DoReMi learns domain weights with a
proxy model and RegMix predicts mixture performance from small training
runs~\citep{doremi,regmix}. These approaches differ in their use of target
examples, gradients, and training feedback. \ours{} provides a standardized
protocol for evaluating
whether frontier models can select useful subsets and forecast their
relative downstream performance. Subsequent training and evaluation
measure both top-subset performance and ranking accuracy.

\section{Setup of DataSense-Bench}
\label{sec:benchmark}
\ours{} aims to evaluate whether a CLI agent can judge the training value of data \emph{before observing training outcomes}.  We explain how we set up our benchmark in detail as follows.


\begin{figure}[!t]
\centering
\includegraphics[width=\linewidth]{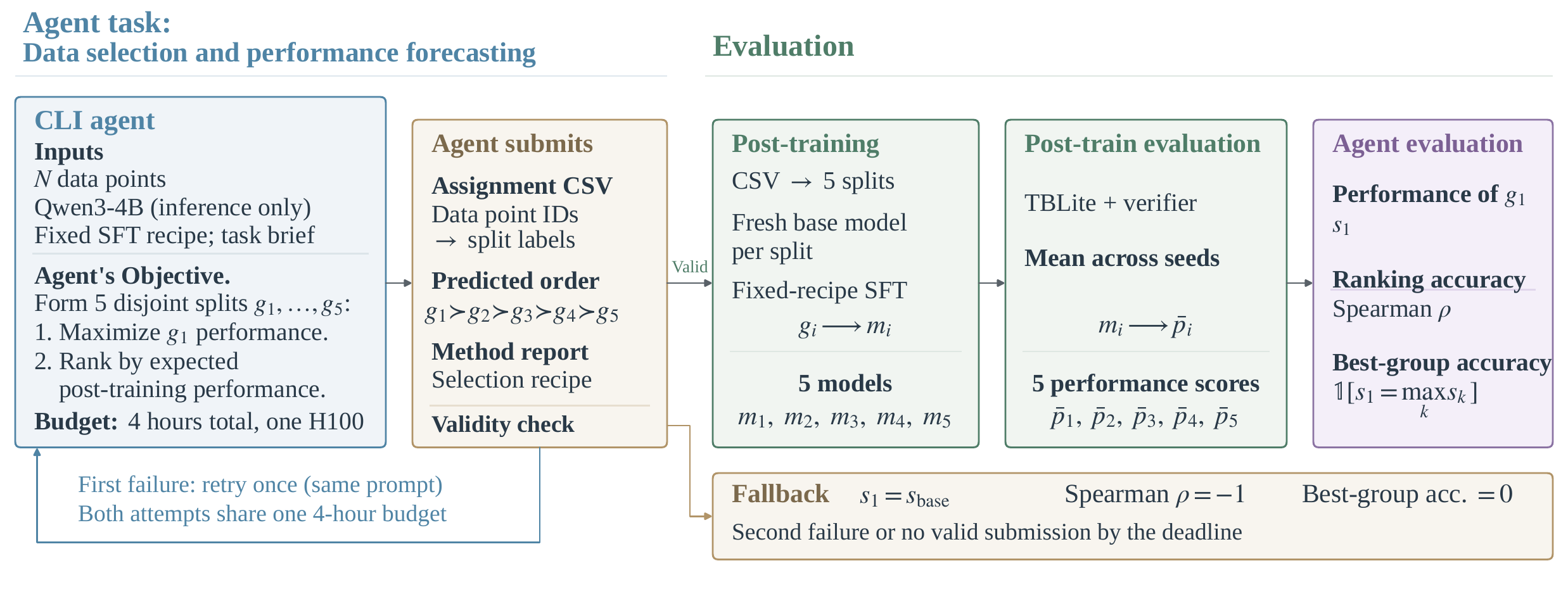}
\vspace{-10pt}
\caption{\textbf{From data-selection predictions to post-hoc verification.}
Illustrated for terminal problem solving, the CLI agent analyzes candidate trajectories with inference-only access to
the base model. It delivers a row-to-group assignment CSV and a method report
specifying its selection recipe. We validate five disjoint groups of the specified size and freeze their membership and predicted order. Only then does
evaluation perform SFT and downstream scoring, testing selection quality,
best-group accuracy, and agreement with the full predicted ordering.}
\label{fig:pipeline}
\vspace{-10pt}
\end{figure}

\subsection{Task for CLI Agents: Select Useful Data and Predict Its Value}
\label{sec:experimental-setup}
Let $D$ be a pool of $N$ candidate trajectories, $M$ a base model, and $R$ the
fine-tuning procedure used to test a submission. The agent selects $K$
subsets $G=(g_1,\ldots,g_K)$ with $g_k\subseteq D$, $|g_k|=n$, and
$g_i\cap g_j=\varnothing$ for $i\ne j$. 
The group labels encode its prediction: training on $g_1$ should give the
highest downstream score, followed by $g_2$, through $g_K$.  Figure~\ref{fig:selection-prompt} provides a
condensed prompt example used in our benchmark, and the full prompt appears in Appendix~\ref{app:prompt}.

\textbf{Post-training Data.} We study two tasks that require learning from
multi-step interactions: terminal problem solving and multi-turn tool use.
Both expose intermediate actions and feedback, allowing agents to assess
which trajectories may provide useful supervision. For each task, agents
select five disjoint groups of complete trajectories, and we independently
fine-tune Qwen3-4B~\footnote{\url{https://huggingface.co/Qwen/Qwen3-4B}}
on each group using supervised fine-tuning (SFT). Whole trajectories are
selected before the training procedure converts them into training samples.

\emph{Tool usage.} The training data come from the EnvScaler SFT
pool~\citep{envscaler}, with validation trajectories excluded. These
trajectories involve selecting functions, supplying arguments, and
responding to tool feedback. Each selected group contains 50 complete
trajectories. 
We evaluate the fine-tuned models on BFCL
V3\footnote{\url{https://gorilla.cs.berkeley.edu/blogs/13_bfcl_v3_multi_turn.html}}, measuring multi-turn tool-use performance.

\emph{Terminal problem solving.} The training pool is
OpenThoughts-Agent-SFT-10K~\citep{otagent}\footnote{\raggedright
Training data: \href{https://huggingface.co/datasets/open-thoughts/OpenThoughts-Agent-SFT-10K}{\texttt{open-thoughts/OpenThoughts-Agent-SFT-10K}}}, which contains
trajectories pairing task instructions with terminal interactions,
including commands, their outputs, and error recovery.
Each selected group contains 1,000 trajectories. We evaluate the fine-tuned
models on TBLite~\citep{otagent}, where automated task verifiers
measure success on terminal tasks.

We will explain BFCL V3 and TBLite later, and the agents are not allowed to access these downstream evaluation tasks. 

\textbf{Agent's Objectives.} We ask agents to select five groups and specify two objectives, as shown in Figure~\ref{fig:selection-prompt}:
\begin{itemize}[nosep, leftmargin=2em]
    \item to make \ensuremath{g_1} as strong as possible, ideally outperforming random selection at the same size.
    \item to order the groups by expected post-training performance.
\end{itemize} 
These objectives assess complementary abilities: ranking evaluates the ordering of weak groups, whereas selection evaluates the ability to identify a useful subset.

\textbf{Resources that the agent can access.}
The agent receives the task prompt, candidate data, base model, task configuration, and a supervised fine-tuning script. The prompt specifies the downstream task types without revealing evaluation instances, while leaving the choice and combination of selection signals to the agent. Each selection run has inference-only access to the base model and is allocated one H100 80\,GB GPU and four hours to analyze the data and submit a selection. Subsequent training and evaluation are performed outside this time budget. Within each task, all agents receive the same fixed training recipe, which is used to fine-tune every selected group. We evaluate ten CLI agents on TBLite and six on BFCL, with three independent selection runs per agent and task. Please refer to Table~\ref{tab:instances} in Appendix~\ref{app:configurations} for more experimental configuration.

\begin{figure}[t]
\begin{tcolorbox}[colback=blue!2,colframe=blue!25,boxrule=0.4pt,
  arc=1mm,left=6pt,right=6pt,top=5pt,bottom=5pt,
  title={Example selection prompt (condensed)},fonttitle=\bfseries,
  coltitle=black,colbacktitle=blue!8,fontupper=\small]
\textbf{Task.} Select training data for Qwen3-4B from a pool of 10,000 terminal
trajectories. Read \texttt{config.yaml} for the data and model paths,
sequence-length limit, and group sizes. Inspect the candidate conversations
and \texttt{recipe/sft.py} to understand the SFT procedure, including
formatting, truncation, and assistant-token supervision.

\smallskip
\textbf{Objectives.} Produce five disjoint groups of 1,000 complete trajectories each. Make \ensuremath{g_1}
as strong as possible, ideally better than a random subset of the same
size, and rank \ensuremath{g_1}--\ensuremath{g_5} by expected post-training performance. Each selected
subset will be used to fine-tune a fresh model, whose terminal-task
performance will be checked by automated tests.

\smallskip
\textbf{Resources and limits.} Choose your own selection method. You may inspect data, execute analysis
code, and run base-model forward passes. You have one H100 80\,GB GPU and
four hours. Do not train models, update weights, modify or execute the SFT
script, or access evaluation tasks and tests.

\smallskip
\textbf{Deliverables.} Submit \texttt{outputs/group\_assignment.csv} with columns
\texttt{index,group\_id} and \texttt{outputs/method.md} explaining your
recipe. Check valid row indices, exact group sizes, and no overlap before
finishing. Your groups and predicted order must be fixed before seeing
training or evaluation outcomes.
\end{tcolorbox}
\caption{Condensed terminal-task prompt specifying the selection and ranking objectives,
permitted tools, computation budget, and required deliverables. The prompt for BFCL is described in Appendix~\ref{app:prompt-bfcl}.}
\label{fig:selection-prompt}
\end{figure}

\textbf{Submission Protocol.}
Each agent is asked to submit a group-assignment CSV and a short method
report explaining the selection recipe and its rationale. The CSV assigns
valid candidate-pool indices to five disjoint groups: 50 for BFCL and 1,000 complete trajectories
per group for TBLite, omitting unselected rows. Group labels \(g_1\) through \(g_5\) indicate the predicted ranking by expected post-training performance, from highest to lowest.
The agent must write and verify both files within the selection time budget, finalizing group composition and ordering before observing any training or evaluation outcomes; no trained model is submitted.

\subsection{Post-check and Post-training Evaluation}
\label{sec:execution-protocol}
After the agent submits its selection, the benchmark validates the
submission and tests its predicted ordering, shown in Figure~\ref{fig:pipeline}.

\textbf{Submission validation.} We check the assignment against the
submission protocol in Section~\ref{sec:experimental-setup} and verify
that the method report is present. For BFCL, we also cross-check row
positions against source trajectory identifiers and exclude validation
trajectories. CLI completion and valid deliverables
are checked separately.
If the first attempt fails to produce a valid submission, the agent may
retry once using the same prompt within the remaining four-hour selection
budget. The two attempts share this budget. If the second attempt also
fails, or no valid submission is produced before the budget expires, we
apply the failure scores defined in Section~\ref{sec:evaluation-metrics}.

\textbf{Post-training evaluation.} Each group is used to fine-tune a fresh
copy of the base model under its task-specific training procedure. Within
each task, all resulting models use the same evaluation suite and scoring protocol.
Let $s_k$ denote the downstream score (TBLite average reward or BFCL accuracy) for the model trained on $g_k$, averaged
across evaluation tasks and evaluation seeds. These outcomes are used only to
assess the submitted selection and ordering, not to revise them.
Section~\ref{sec:tblite-evaluation} describes both downstream evaluation protocols.

\subsection{Evaluation Metrics}
\label{sec:evaluation-metrics}
We want to evaluate two complementary capabilities of the data selection agents: whether the agent selects useful training data, and whether it can anticipate which data choices will be more useful. We therefore evaluate the performance of its
first-ranked group and the agreement between its predicted and observed
ordering.
\textbf{Selection quality.} The score $s_1$ measures the utility of the
data the agent recommends most strongly. To assess whether this choice
improves on an uninformed selection, we also report its gain over random
selection, $\Delta_{\mathrm{random}}=s_1-s_{\mathrm{random}}$, using the same group size and training procedure. A positive gain
supports the selection objective, but does not establish that the agent
can compare its five groups.
\textbf{Ranking accuracy.} We test that comparison at two levels.
Best-group accuracy asks whether the agent's first choice is actually the
best among its selected groups:
$A_{\mathrm{best}}=\mathbb{1}[s_1=\max_{k=1,\ldots,5}s_k]$.
Spearman correlation~\citep{spearman1904} tests the full predicted ordering:
\(
 S_{\mathrm{rank}}=\rho_{\mathrm{Spearman}}
 \big((4,3,2,1,0),(s_1,\ldots,s_5)\big)
\).
These measures distinguish identifying the best group from correctly
ordering all alternatives. We reverse the group-index scale so that perfect
rank agreement gives $+1$, whereas a reversed ordering gives $-1$.

For valid submissions, we compute both ranking metrics within each selection
run after evaluating all five groups. Best-group accuracy counts a tie for
the highest unrounded score as correct. Spearman correlation uses average
ranks for ties and is undefined when all five outcomes are equal.
Such runs are excluded only from the mean Spearman correlation; they remain
in the averages of selection scores and best-group accuracy.
For failed submissions under Section~\ref{sec:execution-protocol}, we set
$s_1=s_{\mathrm{base}}$, $S_{\mathrm{rank}}=-1$, and $A_{\mathrm{best}}=0$.
These are failure penalties, not correlations computed from tied base-model
scores. Failed runs remain in the averages across selection runs;
$\Delta_{\mathrm{random}}$ uses the fallback $s_1$.

\section{Experiments}
\label{sec:results}
Our experiments ask whether agents select data that improves downstream
performance, whether their predicted group order is reliable, and how they
construct these judgments. We report outcomes first, then analyze the
selection procedures and their resource requirements.

\subsection{Downstream Evaluation}
\label{sec:tblite-evaluation}
We evaluate the fine-tuned models on the two downstream tasks introduced
in Section~\ref{sec:experimental-setup}. The data-selection agents do not
perform these evaluations or access their test instances during the data selection.

\noindent\textbf{TBLite.} OpenThoughts-TBLite contains 100 terminal tasks
across nine categories, including software engineering, debugging, data
processing, and security~\citep{otagent}. We use Harbor's\footnote{\url{https://www.harborframework.com/}}
Terminus-2 scaffold to execute model-generated commands and return terminal
outputs. Each attempt starts in a fresh Daytona\footnote{\url{https://www.daytona.io/docs/en/}}
sandbox and ends when the model finishes or reaches the
task's time limit. Task-supplied verifiers check the resulting environment
or artifacts and assign rewards, including fractional credit where
supported. We report average reward over 100 tasks with three attempts per
task, giving 300 equally weighted scores per checkpoint.

\noindent\textbf{BFCL.} For models trained on EnvScaler trajectories, we
use BFCL V3: 200 cases each for
base interactions, missing functions, missing parameters, and long
context. The evaluator checks tool-use outcomes against the required
state and execution behavior. Each evaluation seed covers all 800
cases; its overall accuracy averages the four equally sized categories.
We use three evaluation seeds per checkpoint.

\subsection{Selection Quality and Ranking Accuracy}
\begin{table}[!t]
\centering\small
\setlength{\tabcolsep}{1.6pt}\setlength{\fboxsep}{0.8pt}\renewcommand{\arraystretch}{0.88}
\caption{Selection performance, best-group accuracy, and rank performance on BFCL and TBLite. Agent-group scores ($s_1$-$s_5$) report the post-training performance, accuracy  (\%) for BFCL, and reward for TBLite (scale from 0-1 to 0-100). We report the average performance across three evaluation seeds and three data selection runs, standard deviation (subscript), and maximum score among different data selection runs. Shading compares group averages within a row; red outlines mark each task's largest maximum among agent-selected groups. $g_1$ gain $\Delta_{\text{random}}$ is relative to random selection. Best-group accuracy $A_{\mathrm{best}}$ measures how often the agent identifies its best group; rank performance $S_{\mathrm{rank}}$ measures Spearman correlation between predicted and observed group orderings (Section~\ref{sec:evaluation-metrics}). More detailed experimental results are in Appendix~\ref{app:eval}.}
\label{tab:main-results}\label{tab:terminal-main}
\resizebox{\linewidth}{!}{\begin{tabular}{@{}clrr@{\hspace{5pt}}rrrr@{\hspace{5pt}}rr@{}}\toprule
Rank & Agent / reference & $s_1$ & \begin{tabular}[c]{@{}c@{}}$g_1$ gain\\[-2pt]\scriptsize $\Delta_{\mathrm{random}}$ \end{tabular} & $s_2$ & $s_3$ & $s_4$ & $s_5$ & $S_{\mathrm{rank}}$ & $A_{\mathrm{best}}$ (\%) \\\midrule
\multicolumn{10}{c}{\textbf{BFCL accuracy (50 trajectories per group)}\quad Base model: 31.67\%}\\\midrule
\midrule
1 & DeepSeek V4 Pro & \colorbox{green!22}{\shortstack{$36.53_{\text{\scriptsize $\pm$1.83}}$\\[-2pt]{\scriptsize {\color{red}\fbox{\color{black}Max: 38.62}}}}} & 1.78 & \colorbox{green!28}{\shortstack{$\mathbf{36.61}_{\text{\scriptsize $\pm$0.72}}$\\[-2pt]{\scriptsize Max: 37.42}}} & \colorbox{green!10}{\shortstack{$35.74_{\text{\scriptsize $\pm$1.12}}$\\[-2pt]{\scriptsize Max: 37.00}}} & \colorbox{green!16}{\shortstack{$36.08_{\text{\scriptsize $\pm$1.44}}$\\[-2pt]{\scriptsize Max: 37.75}}} & \colorbox{green!4}{\shortstack{$34.33_{\text{\scriptsize $\pm$1.16}}$\\[-2pt]{\scriptsize Max: 35.67}}} & 0.65 & 33.33 \\
2 & GPT-6 Astra & \colorbox{green!28}{\shortstack{$\mathbf{36.49}_{\text{\scriptsize $\pm$0.23}}$\\[-2pt]{\scriptsize Max: 36.75}}} & 1.74 & \colorbox{green!22}{\shortstack{$35.31_{\text{\scriptsize $\pm$0.94}}$\\[-2pt]{\scriptsize Max: 36.38}}} & \colorbox{green!10}{\shortstack{$34.74_{\text{\scriptsize $\pm$1.50}}$\\[-2pt]{\scriptsize Max: 36.21}}} & \colorbox{green!16}{\shortstack{$34.93_{\text{\scriptsize $\pm$0.42}}$\\[-2pt]{\scriptsize Max: 35.42}}} & \colorbox{green!4}{\shortstack{$34.58_{\text{\scriptsize $\pm$0.07}}$\\[-2pt]{\scriptsize Max: 34.62}}} & 0.83 & 100.00 \\
3 & Kimi K3 & \colorbox{green!22}{\shortstack{$36.29_{\text{\scriptsize $\pm$0.91}}$\\[-2pt]{\scriptsize Max: 36.92}}} & 1.54 & \colorbox{green!16}{\shortstack{$36.21_{\text{\scriptsize $\pm$0.65}}$\\[-2pt]{\scriptsize Max: 36.92}}} & \colorbox{green!10}{\shortstack{$36.18_{\text{\scriptsize $\pm$1.22}}$\\[-2pt]{\scriptsize Max: 37.58}}} & \colorbox{green!4}{\shortstack{$35.85_{\text{\scriptsize $\pm$1.14}}$\\[-2pt]{\scriptsize Max: 36.62}}} & \colorbox{green!28}{\shortstack{$\mathbf{36.46}_{\text{\scriptsize $\pm$1.27}}$\\[-2pt]{\scriptsize Max: 37.75}}} & 0.02 & 33.33 \\
4 & Opus 5 & \colorbox{green!22}{\shortstack{$35.94_{\text{\scriptsize $\pm$0.72}}$\\[-2pt]{\scriptsize Max: 36.75}}} & 1.19 & \colorbox{green!28}{\shortstack{$\mathbf{36.35}_{\text{\scriptsize $\pm$1.09}}$\\[-2pt]{\scriptsize Max: 37.38}}} & \colorbox{green!16}{\shortstack{$35.35_{\text{\scriptsize $\pm$0.53}}$\\[-2pt]{\scriptsize Max: 35.96}}} & \colorbox{green!10}{\shortstack{$34.29_{\text{\scriptsize $\pm$2.29}}$\\[-2pt]{\scriptsize Max: 36.54}}} & \colorbox{green!4}{\shortstack{$31.08_{\text{\scriptsize $\pm$3.45}}$\\[-2pt]{\scriptsize Max: 35.04}}} & 0.77 & 0.00 \\
5 & Fable 5.1 & \colorbox{green!22}{\shortstack{$35.21_{\text{\scriptsize $\pm$0.94}}$\\[-2pt]{\scriptsize Max: 36.00}}} & 0.46 & \colorbox{green!16}{\shortstack{$34.96_{\text{\scriptsize $\pm$1.67}}$\\[-2pt]{\scriptsize Max: 36.58}}} & \colorbox{green!10}{\shortstack{$34.87_{\text{\scriptsize $\pm$0.07}}$\\[-2pt]{\scriptsize Max: 34.96}}} & \colorbox{green!4}{\shortstack{$34.10_{\text{\scriptsize $\pm$0.24}}$\\[-2pt]{\scriptsize Max: 34.38}}} & \colorbox{green!28}{\shortstack{$\mathbf{35.57}_{\text{\scriptsize $\pm$1.53}}$\\[-2pt]{\scriptsize Max: 37.17}}} & 0.23 & 33.33 \\
6 & Gemini 3.5 Flash & \colorbox{green!22}{\shortstack{$35.10_{\text{\scriptsize $\pm$1.56}}$\\[-2pt]{\scriptsize Max: 36.42}}} & 0.35 & \colorbox{green!16}{\shortstack{$34.18_{\text{\scriptsize $\pm$1.10}}$\\[-2pt]{\scriptsize Max: 35.29}}} & \colorbox{green!10}{\shortstack{$33.57_{\text{\scriptsize $\pm$0.49}}$\\[-2pt]{\scriptsize Max: 34.12}}} & \colorbox{green!4}{\shortstack{$33.07_{\text{\scriptsize $\pm$3.35}}$\\[-2pt]{\scriptsize Max: 35.21}}} & \colorbox{green!28}{\shortstack{$\mathbf{35.60}_{\text{\scriptsize $\pm$0.94}}$\\[-2pt]{\scriptsize Max: 36.33}}} & -0.27 & 33.33 \\
\midrule
 & Random selection & $34.75_{\pm1.19}$ & 0.00 & \multicolumn{4}{c}{--} & -- & -- \\
\midrule
\midrule
\multicolumn{10}{c}{\textbf{TBLite average reward (1,000 trajectories per group)}\quad Base model: 4.05\%}\\\midrule
\midrule
1 & DeepSeek V4 Pro & \colorbox{blue!28}{\shortstack{$\mathbf{12.54}_{\text{\scriptsize $\pm$1.17}}$\\[-2pt]{\scriptsize Max: 13.46}}} & 0.99 & \colorbox{blue!10}{\shortstack{$11.75_{\text{\scriptsize $\pm$1.76}}$\\[-2pt]{\scriptsize Max: 13.26}}} & \colorbox{blue!4}{\shortstack{$11.65_{\text{\scriptsize $\pm$1.78}}$\\[-2pt]{\scriptsize Max: 13.69}}} & \colorbox{blue!22}{\shortstack{$12.19_{\text{\scriptsize $\pm$0.25}}$\\[-2pt]{\scriptsize Max: 12.47}}} & \colorbox{blue!16}{\shortstack{$12.11_{\text{\scriptsize $\pm$1.88}}$\\[-2pt]{\scriptsize Max: 13.21}}} & 0.13 & 66.67 \\
2 & Fable 5.1 & \colorbox{blue!28}{\shortstack{$\mathbf{12.50}_{\text{\scriptsize $\pm$2.05}}$\\[-2pt]{\scriptsize Max: 14.83}}} & 0.95 & \colorbox{blue!10}{\shortstack{$11.20_{\text{\scriptsize $\pm$0.98}}$\\[-2pt]{\scriptsize Max: 12.24}}} & \colorbox{blue!22}{\shortstack{$12.01_{\text{\scriptsize $\pm$0.92}}$\\[-2pt]{\scriptsize Max: 12.87}}} & \colorbox{blue!4}{\shortstack{$10.22_{\text{\scriptsize $\pm$0.84}}$\\[-2pt]{\scriptsize Max: 11.18}}} & \colorbox{blue!16}{\shortstack{$11.62_{\text{\scriptsize $\pm$1.57}}$\\[-2pt]{\scriptsize Max: 13.22}}} & 0.30 & 33.33 \\
3 & Gemini 3.5 Flash & \colorbox{blue!16}{\shortstack{$12.07_{\text{\scriptsize $\pm$0.79}}$\\[-2pt]{\scriptsize Max: 12.98}}} & 0.52 & \colorbox{blue!28}{\shortstack{$\mathbf{13.42}_{\text{\scriptsize $\pm$0.66}}$\\[-2pt]{\scriptsize Max: 13.93}}} & \colorbox{blue!10}{\shortstack{$11.82_{\text{\scriptsize $\pm$1.56}}$\\[-2pt]{\scriptsize Max: 13.25}}} & \colorbox{blue!22}{\shortstack{$12.58_{\text{\scriptsize $\pm$0.98}}$\\[-2pt]{\scriptsize Max: 13.70}}} & \colorbox{blue!4}{\shortstack{$10.96_{\text{\scriptsize $\pm$2.69}}$\\[-2pt]{\scriptsize Max: 13.55}}} & 0.30 & 0.00 \\
4 & GPT-6 Astra & \colorbox{blue!22}{\shortstack{$11.95_{\text{\scriptsize $\pm$1.96}}$\\[-2pt]{\scriptsize Max: 14.18}}} & 0.40 & \colorbox{blue!28}{\shortstack{$\mathbf{13.35}_{\text{\scriptsize $\pm$1.86}}$\\[-2pt]{\scriptsize Max: 14.52}}} & \colorbox{blue!10}{\shortstack{$10.60_{\text{\scriptsize $\pm$0.81}}$\\[-2pt]{\scriptsize Max: 11.31}}} & \colorbox{blue!16}{\shortstack{$11.60_{\text{\scriptsize $\pm$1.88}}$\\[-2pt]{\scriptsize Max: 13.56}}} & \colorbox{blue!4}{\shortstack{$10.44_{\text{\scriptsize $\pm$2.29}}$\\[-2pt]{\scriptsize Max: 12.11}}} & 0.17 & 33.33 \\
5 & Kimi K3 & \colorbox{blue!22}{\shortstack{$11.94_{\text{\scriptsize $\pm$1.46}}$\\[-2pt]{\scriptsize Max: 13.41}}} & 0.38 & \colorbox{blue!10}{\shortstack{$11.02_{\text{\scriptsize $\pm$2.00}}$\\[-2pt]{\scriptsize Max: 13.09}}} & \colorbox{blue!16}{\shortstack{$11.34_{\text{\scriptsize $\pm$0.63}}$\\[-2pt]{\scriptsize Max: 12.04}}} & \colorbox{blue!28}{\shortstack{$\mathbf{13.05}_{\text{\scriptsize $\pm$2.09}}$\\[-2pt]{\scriptsize {\color{red}\fbox{\color{black}Max: 15.16}}}}} & \colorbox{blue!4}{\shortstack{$10.81_{\text{\scriptsize $\pm$0.57}}$\\[-2pt]{\scriptsize Max: 11.42}}} & 0.07 & 33.33 \\
6 & GPT-5.5 & \colorbox{blue!22}{\shortstack{$11.87_{\text{\scriptsize $\pm$1.04}}$\\[-2pt]{\scriptsize Max: 12.80}}} & 0.31 & \colorbox{blue!4}{\shortstack{$9.81_{\text{\scriptsize $\pm$1.10}}$\\[-2pt]{\scriptsize Max: 10.56}}} & \colorbox{blue!16}{\shortstack{$11.70_{\text{\scriptsize $\pm$1.39}}$\\[-2pt]{\scriptsize Max: 13.31}}} & \colorbox{blue!10}{\shortstack{$11.54_{\text{\scriptsize $\pm$1.49}}$\\[-2pt]{\scriptsize Max: 12.74}}} & \colorbox{blue!28}{\shortstack{$\mathbf{12.45}_{\text{\scriptsize $\pm$0.25}}$\\[-2pt]{\scriptsize Max: 12.64}}} & -0.23 & 0.00 \\
7 & Opus 5 & \colorbox{blue!16}{\shortstack{$11.60_{\text{\scriptsize $\pm$0.85}}$\\[-2pt]{\scriptsize Max: 12.51}}} & 0.05 & \colorbox{blue!28}{\shortstack{$\mathbf{12.50}_{\text{\scriptsize $\pm$1.33}}$\\[-2pt]{\scriptsize Max: 13.73}}} & \colorbox{blue!22}{\shortstack{$11.94_{\text{\scriptsize $\pm$2.05}}$\\[-2pt]{\scriptsize Max: 13.70}}} & \colorbox{blue!10}{\shortstack{$10.94_{\text{\scriptsize $\pm$2.26}}$\\[-2pt]{\scriptsize Max: 13.53}}} & \colorbox{blue!4}{\shortstack{$10.35_{\text{\scriptsize $\pm$1.99}}$\\[-2pt]{\scriptsize Max: 11.71}}} & 0.30 & 0.00 \\
8 & Sonnet 5 & \colorbox{blue!10}{\shortstack{$11.55_{\text{\scriptsize $\pm$1.57}}$\\[-2pt]{\scriptsize Max: 13.00}}} & 0.00 & \colorbox{blue!22}{\shortstack{$11.95_{\text{\scriptsize $\pm$0.78}}$\\[-2pt]{\scriptsize Max: 12.48}}} & \colorbox{blue!4}{\shortstack{$10.67_{\text{\scriptsize $\pm$0.79}}$\\[-2pt]{\scriptsize Max: 11.57}}} & \colorbox{blue!16}{\shortstack{$11.82_{\text{\scriptsize $\pm$0.57}}$\\[-2pt]{\scriptsize Max: 12.43}}} & \colorbox{blue!28}{\shortstack{$\mathbf{12.94}_{\text{\scriptsize $\pm$1.77}}$\\[-2pt]{\scriptsize Max: 14.92}}} & -0.20 & 33.33 \\
\midrule
 & Random selection & {\shortstack{$11.55_{\text{\scriptsize $\pm$1.04}}$}} & 0.00 & \multicolumn{4}{c}{--} & -- & -- \\
\midrule
9 & GPT-5.6 Sol & \colorbox{blue!10}{\shortstack{$11.26_{\text{\scriptsize $\pm$1.50}}$\\[-2pt]{\scriptsize Max: 12.96}}} & -0.29 & \colorbox{blue!4}{\shortstack{$11.22_{\text{\scriptsize $\pm$0.50}}$\\[-2pt]{\scriptsize Max: 11.77}}} & \colorbox{blue!16}{\shortstack{$11.39_{\text{\scriptsize $\pm$1.12}}$\\[-2pt]{\scriptsize Max: 12.63}}} & \colorbox{blue!28}{\shortstack{$\mathbf{12.90}_{\text{\scriptsize $\pm$0.50}}$\\[-2pt]{\scriptsize Max: 13.29}}} & \colorbox{blue!22}{\shortstack{$12.75_{\text{\scriptsize $\pm$2.50}}$\\[-2pt]{\scriptsize Max: 14.78}}} & -0.50 & 0.00 \\
10 & Fable 5 & \colorbox{blue!16}{\shortstack{$11.00_{\text{\scriptsize $\pm$1.08}}$\\[-2pt]{\scriptsize Max: 12.14}}} & -0.55 & \colorbox{blue!4}{\shortstack{$10.53_{\text{\scriptsize $\pm$1.86}}$\\[-2pt]{\scriptsize Max: 12.08}}} & \colorbox{blue!28}{\shortstack{$\mathbf{13.40}_{\text{\scriptsize $\pm$1.20}}$\\[-2pt]{\scriptsize Max: 14.52}}} & \colorbox{blue!22}{\shortstack{$11.50_{\text{\scriptsize $\pm$1.74}}$\\[-2pt]{\scriptsize Max: 13.44}}} & \colorbox{blue!10}{\shortstack{$10.92_{\text{\scriptsize $\pm$1.14}}$\\[-2pt]{\scriptsize Max: 12.14}}} & -0.03 & 0.00 \\
\bottomrule\end{tabular}}\end{table}

\label{sec:terminal-results}
Table~\ref{tab:main-results} separates two questions: whether agents
choose useful training data and whether they correctly rank the groups
they select. 

\textbf{Selection yields limited gains over random selection on both tasks.}
On BFCL, all six agents' average $g_1$ accuracies exceed random
selection's 34.75\%, but gains range from only 0.35 to 1.78 \%
points. DeepSeek V4 Pro has the highest average at 36.53\%. Its run-3
$g_1$ achieves the highest accuracy among agent-selected checkpoints, 38.62\%.
On TBLite, average $g_1$ rewards range from 11.00\% to 12.54\%, compared
with 11.55\% for random selection. Two of ten agents underperform this
reference, and DeepSeek V4 Pro's leading score improves on it by only
0.99 points. Thus, elaborate selection recipes yield modest gains on
both tasks, with less consistent improvements on TBLite. 

\textbf{Selecting useful data does not imply reliably ranking the selected groups.}
BFCL reveals a gap between selection quality and ranking accuracy:
all six agents improve over Random with $g_1$, yet their average
Spearman correlations range from $-0.27$ to $0.83$. GPT-6 Astra
identifies the best group in all three runs and achieves a correlation
of 0.83. In contrast, Kimi K3 achieves 36.29\% average $g_1$
accuracy but a correlation of only 0.02, while Gemini's negative
correlation indicates disagreement with the predicted ordering.
Conversely, Opus has a correlation of 0.77 but never identifies the best
group, illustrating that agreement over all five positions can coexist
with an incorrect first choice.
The difficulty is more widespread on TBLite, where correlations range
from $-0.50$ to $0.30$. Even DeepSeek V4 Pro, the strongest agent by
average $g_1$ reward, achieves only 0.13. Kimi K3 places the
highest-performing agent-selected subset fourth: its run-3 $g_4$ reaches 15.16\%
average reward. Together, the two tasks show that gains over Random
do not establish an ability to compare alternative data choices;
no agent identifies the best group in all three runs on both tasks.
Astra, for example, does so in three of three BFCL runs but only one of
three TBLite runs. Appendix
Tables~\ref{tab:terminal-orders} and~\ref{tab:bfcl-orders} report the
individual selection runs.

\begin{figure}[t]
\centering
\begin{tcolorbox}[colframe=brown!60!black,colback=brown!3,boxrule=.4pt,arc=0pt,left=6pt,right=6pt,top=5pt,bottom=5pt]
\small\raggedright
\renewcommand{\arraystretch}{1.12}
\begin{tabular}{@{}>{\raggedright\arraybackslash}p{.12\linewidth}>{\raggedright\arraybackslash}p{.82\linewidth}@{}}
\toprule
Time & Executed commands and resulting decisions \\\midrule
00:00:02 & \texttt{cat config.yaml}; \texttt{cat recipe/sft.py}; \texttt{nvidia-smi}. Read the selection configuration and SFT recipe; inspect available compute.\\[3pt]
00:00:49--00:03:13 & \texttt{python - <{}<'EOF'}: read \texttt{data/train.parquet}, print example turns and task descriptions, and count completion markers, parser warnings, and response formats.\\[3pt]
00:04:04 & Write and run \texttt{python work/features.py}. Extract structural features and token counts into \texttt{features.parquet}; save task descriptions in \texttt{tasks.parquet}.\\[3pt]
00:06:44--00:08:43 & Write \texttt{work/score\_gpu.py}; pilot with \texttt{timeout 900 python work/score\_test.py}. Revise scoring priority, then launch \texttt{python work/score\_gpu.py > work/gpu\_log.txt 2>\&1}: frozen-model assistant-token loss, up to 32,768 tokens.\\[3pt]
00:10:24 & Inline Python computes task duplicates and TF-IDF similarity, fits \texttt{MiniBatchKMeans(n\_clusters=50)}, and saves \texttt{work/features2.parquet}.\\[3pt]
00:12:01--00:14:35 & Write and repeatedly run \texttt{python work/select.py}. Inspect the provisional groups; replace fixed cluster caps with pool-relative caps, add an 8\% Tezos cap, then add source caps. These changes precede downstream evaluation.\\[3pt]
00:40:43 & Inline Python reads \texttt{work/gpu\_scores.pkl} and checks loss distributions, source-level statistics, correlations, and extreme-loss examples while GPU scoring continues.\\[3pt]
02:04:37 & Rerun \texttt{python work/select.py}; inspect the CSV with \texttt{head -3} and \texttt{wc -l}. Inline assertions check column names, unique in-range indices, group IDs, and exactly 1,000 rows per group.\\[3pt]
02:05:18 & \texttt{cat >> outputs/method.md <{}<'EOF'}: append final group statistics and compute details; list the delivered files in \texttt{outputs/}.\\
\bottomrule
\end{tabular}\par\medskip
\textbf{Final recipe.}
\textcolor{orange!80!black}{\textit{Score.}} Combine completion ($+2$), usable assistant-token supervision,
command substance, and writing/testing cues with soft penalties for format
faults, context handoff, truncation, empty or repetitive actions, and narrow
domain coverage. Add loss-outlier penalties for both unusually high and low
average loss, high-loss token fractions, and extreme last/max-turn loss;
loss is not a monotonic preference for harder or easier examples.
\textcolor{orange!80!black}{\textit{Construct groups.}} Sort by composite score. For $g_1$, greedily take
1,000 rows with one trajectory per task and cap cluster $c$ at
$\max(12,\lceil0.20n_c\rceil)$, where $n_c$ is its pool size. Repeat on
unassigned rows for $g_2$, using $0.25n_c$. Both groups cap software-engineering,
synthetic, and other sources at 40\%, 45\%, and 35\%, respectively, and Tezos
at 8\%. Assign the next two highest-scoring blocks to $g_3$ and $g_4$ without
these diversity caps; assign the lowest-scoring remaining 1,000 rows to
$g_5$. Deliver five disjoint groups in \texttt{group\_assignment.csv} and
explain the procedure in \texttt{method.md}.
\end{tcolorbox}
\vspace{-10pt}
\caption{Fable 5.1's data-selection workflow, from inspection to final recipe.}
\vspace{-20pt}
\label{fig:case-traces}
\end{figure}

\subsection{How Agents Make Selection Decisions}
\label{sec:behavior}
We analyze 30 TBLite and 18 BFCL selection procedures and identify three patterns. \textbf{Common signals, different value judgments.}
Across TBLite recipes, completion cues appear in 30 procedures, length
features in 29, format checks in 28, base-model loss in 23, and diversity
controls in 21. Yet agents interpret these signals differently. One
DeepSeek V4 Pro recipe rewards higher first-assistant-turn loss after
penalizing formatting defects:
\(q_{\mathrm{DS}}=\ell(1-0.10e-0.05w-0.50j)(0.5+0.5c)\),
where $e,w,j$ are extra-text, warning, and invalid-JSON rates, and $c$
indicates completion. Kimi K3 instead favors lower length-normalized loss:
\(q_{\mathrm{Kimi}}=-z_{\mathrm{length}}(\ell)+1.2c-2t,\)
where loss is standardized within length deciles and $t$ indicates
truncation. Fable 5.1 uses several loss statistics as anomaly
penalties on a structural score, with task coverage and source caps.
These are different judgments of training
value, although their loss measurements also differ in coverage and normalization. \textbf{Recipes change across independent runs.}
The same agent can use different data-selection rules across runs.
For example, Astra's three TBLite runs assess quality through generated
correctness audits, separate prompted judgments of demonstration quality,
task accomplishment and complexity, and a five-level rating with
execution-evidence checks, respectively. The variation also occurs in
BFCL: Kimi moves from rejecting loss outliers to preferring intermediate
loss and then lower loss. Thus, a single run's recipe does not fully
characterize an agent's selection strategy. \textbf{Agents construct the predicted ordering in different ways.}
The ordering is part of the selection procedure: agents can partition a
scored candidate list or apply different selection criteria to different
groups. Astra's third TBLite run combines these approaches, constructing
$g_1$ with deduplication and skill coverage before drawing the remaining
groups from descending score bands. In BFCL, Fable 5.1's first run places
its lowest-scoring candidates in $g_5$, whereas Gemini's second run
explicitly assigns looping and error-heavy trajectories to that group.
These procedures encode the agents' predictions of relative training
value; Section~\ref{sec:terminal-results} evaluates how well those predictions
match post-training performance.

\subsection{Case Study}
Figure~\ref{fig:case-traces} shows Fable 5.1's data selection for the terminal task, including data inspection and the final recipe. The agent first builds structural
features and base-model loss statistics, then revises its grouping rules
after inspecting the resulting subsets. Its final recipe uses loss to
penalize outliers.
It also applies task, cluster, and source constraints to $g_1$ and $g_2$,
while constructing later groups from score bands without those diversity
caps. This case shows that the submitted ordering combines example-level
scores with group-level coverage choices. In this run, $g_1$ achieves 14.83\% reward, but
$g_2$ (11.08\%) falls below $g_3$ (12.12\%). The best first choice
therefore coexists with an error in the remaining order. Appendix~\ref{app:accepted-methods} and
Table~\ref{tab:bfcl-method-summary} provide the remaining recipes.

\subsection{Resource Requirements}
The benchmark separates the cost of making a data recommendation from
the cost of testing it through SFT and downstream evaluation.

\noindent\textbf{CLI agent cost and time budget.}
Table~\ref{tab:api-cost} summarizes the agent API cost for data selection. Selection estimates use cumulative token ledgers and exclude
local inference, training, and evaluation charges. Figure~\ref{fig:time-budget} shows  time cost against the four-hour time
budget, including tool execution and waits. Measurement details and pricing
assumptions are in Appendix~\ref{app:runtime}.

\begin{table}[!htbp]
\begin{minipage}[t]{0.56\linewidth}
\vspace{0pt}\centering\scriptsize
\setlength{\tabcolsep}{2pt}\renewcommand{\arraystretch}{1.08}
\caption{Mean selection cost (USD; ranges reflect billing assumptions) and training/evaluation hours per checkpoint (medians; TBLite evaluation estimated). }
\label{tab:api-cost}
\begin{tabular}{@{}lrr@{\hspace{5pt}}lrr@{}}\toprule
Agent & TBLite & BFCL & Agent & TBLite & BFCL\\\midrule
Fable 5.1 & 5.98 & 5.35 & Opus 5 & 3.69 & 4.91\\
GPT-6 Astra & 11.11 & $\geq$9.35 & Kimi K3 & 2.41 & 1.38\\
DeepSeek & 0.28--0.57 & 0.38--0.78 & Gemini & 1.11--1.34 & 0.84--1.01\\
Fable 5 & 5.10 & -- & Sonnet 5 & 3.63 & --\\
GPT-5.5 & 2.54 & -- & GPT-5.6 Sol & 0.70 & --\\\midrule
\multicolumn{6}{@{}l@{}}{Training / evaluation (h): TBLite 2.04 / $\approx$3.93; BFCL 0.29 / 3.30.}\\
\bottomrule\end{tabular}
\end{minipage}\hfill
\begin{minipage}[t]{0.42\linewidth}
\vspace{0pt}\centering
\includegraphics[width=\linewidth]{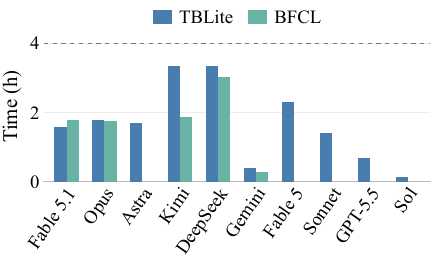}
\end{minipage}
\par
\begin{minipage}[t]{0.56\linewidth}
\vspace{0pt}

\end{minipage}\hfill
\begin{minipage}[t]{0.42\linewidth}
\vspace{-20pt}
\makeatletter\def\@captype{figure}\makeatother
\caption{\textbf{Time cost in data selection.} The dashed line marks the time budget.}
\label{fig:time-budget}
\end{minipage}
\end{table}

\noindent\textbf{Training and evaluation time.}
Evaluation is the larger time component on both tasks
(Table~\ref{tab:api-cost}). Each selection requires five trained
checkpoints. TBLite evaluation time is a scheduling estimate for 300
attempts with 32 concurrent trials on two H100 GPUs, including model
loading; it is not a measured end-to-end runtime. BFCL medians cover 90 training runs and 35 complete checkpoint evaluations,
with all three evaluation seeds sharing one model server on one GPU.

\section{Conclusion}
\label{sec:conclusion}
Anticipating which changes will improve a model can help recursive
self-improvement allocate training resources more effectively. As these systems take on
research roles, data sense becomes a concrete part of that challenge:
connecting properties of training examples to what a model will learn
from them. We introduce \ours{}
to evaluate this data sense through selecting useful subsets and predicting
their relative post-training performance. Across terminal problem solving
and multi-turn tool use, agents construct plausible, data-specific recipes,
but selection gains over random selection are limited. Agents do not
reliably rank their selected groups, and this ability does not hold
consistently across tasks. These results
highlight a gap between producing a convincing analysis of data and making
reliable decisions about its training value, motivating data sense as a
capability to evaluate when assessing progress toward AI scientists and their recursive self-improvement.

\noindent\textbf{Limitations and future work.} Our benchmark studies data sense
through selecting and ranking existing data for SFT. Future work should
extend this scope to reinforcement-learning post-training and the generation
of new training data, testing whether agents can anticipate which experiences
to collect or create and how those choices will improve learning.

\section*{AI Use Statement}
OpenAI Codex assists with drafting data-selection prompts, preparing code snapshots, monitoring experiments, and polishing the manuscript. Frontier models also act as experimental participants, as described in Section~\ref{sec:experimental-setup}.
Numerical results are computed from the recorded selections and downstream
evaluations. The authors are responsible for the final text, methods,
results, and claims.

\section*{Broader Impact}
As AI systems increasingly influence how models are trained, their data choices
can shape whose needs and experiences those models serve. \ours{} makes these
choices testable, supporting closer scrutiny of claims about autonomous AI
research and recursive self-improvement. Better data selection could reduce
wasted training compute, but optimizing benchmark performance alone may favor
narrow capabilities or amplify biases in the candidate data. Strong results
should therefore be considered alongside data provenance, privacy, and
representativeness. Repeatedly selecting data for aggregate performance could exclude rare but
important cases. Human review of these choices remains necessary, especially
when the resulting models are used in settings where errors affect people.
We hope this work supports evidence-based oversight of AI research agents.

\bibliographystyle{iclr2027_conference}
\bibliography{iclr2027_conference}
\clearpage
\appendix
\addtocontents{toc}{\protect\setcounter{tocdepth}{2}}
\section{Experimental Details}
\label{app:configurations}

\subsection{Training Configuration}
\label{app:recipe}
Table~\ref{tab:instances} lists the configurations for both tasks.
We use full-parameter SFT with assistant-token supervision; TBLite
supervision includes thinking, and its gradient-clipping threshold is
$10^{-3}$. Selected BFCL trajectories are converted into
assistant-supervised examples before training. Within each task, all
agent-selected groups share the realized training configuration.

\begin{table}[!htbp]
\centering\small
\caption{Configuration of the two selection-and-ranking tasks. Group sizes count complete trajectories.}
\label{tab:instances}
\setlength{\tabcolsep}{4pt}
\begin{tabular}{lll}\toprule
 & Terminal problem solving & Multi-turn tool use\\\midrule
Training pool & OpenThoughts-Agent 10K & EnvScaler SFT\\
Groups $\times$ group size & $5\times1,000$ & $5\times50$\\
Base model & Qwen3-4B & Qwen3-4B\\
Selectors / runs per selector & 10 / 3 & 6 / 3\\
Selection GPU / time budget & H100 80GB / 4 h & H100 80GB / 4 h\\
Learning rate / epochs & $4\times10^{-5}$ / 7 & $5\times10^{-6}$ / 2\\
Effective batch / training seed & 96 / 42 & 32 / 42\\
Training token cutoff & 32,768 & 16,384\\
Downstream evaluation & TBLite: 100 tasks & BFCL V3: 800 cases\\
Main outcome & Average reward & Overall accuracy\\
Evaluation seeds & 3 & 3\\
\bottomrule\end{tabular}\end{table}

\subsection{Selection Models and Inference Configuration}
\label{app:agent-config}
Table~\ref{tab:agent-identifiers} maps the names in the results to the model identifiers used by the selection agents. Claude Code session headers record version 2.1.270; the retained deployment packages are Codex CLI 0.146.1 and Gemini CLI 0.60.0.
\begin{table}[!htbp]
\centering\small
\caption{Selection-agent model identifiers.}
\label{tab:agent-identifiers}
\setlength{\tabcolsep}{5pt}
\hypersetup{colorlinks=true,urlcolor=blue!70!black}
\begin{tabular}{ll}\toprule
Agent & Model identifier\\\midrule
\href{https://platform.claude.com/docs/en/models/overview}{Fable 5.1} & \texttt{claude-fable-5-1}\\
\href{https://platform.claude.com/docs/en/models/overview}{Opus 5} & \texttt{claude-opus-5}\\
\href{https://developers.openai.com/api/docs/models}{GPT-6 Astra} & \texttt{gpt-6-astra}\\
\href{https://platform.kimi.ai/docs/overview}{Kimi K3} & \texttt{kimi-k3}\\
\href{https://api-docs.deepseek.com/}{DeepSeek V4 Pro} & \texttt{deepseek-v4-pro}\\
\href{https://ai.google.dev/gemini-api/docs/models}{Gemini 3.5 Flash} & \texttt{gemini-3.5-flash}\\
\href{https://platform.claude.com/docs/en/models/overview}{Fable 5} & \texttt{claude-fable-5}\\
\href{https://platform.claude.com/docs/en/models/overview}{Sonnet 5} & \texttt{claude-sonnet-5}\\
\href{https://developers.openai.com/api/docs/models}{GPT-5.5} & \texttt{gpt-5.5}\\
\href{https://developers.openai.com/api/docs/models}{GPT-5.6 Sol} & \texttt{gpt-5.6-sol}\\\bottomrule
\end{tabular}
\end{table}

All selection CLIs run non-interactively with automatic tool approval.
Claude Code uses streaming JSON output and summarized thinking display;
Kimi and DeepSeek use their Anthropic-compatible endpoints through the same CLI.
Codex enables web search and detailed reasoning summaries. The wrappers do not
set a selection temperature, sampling seed, or reasoning-effort override;
these use the CLI/provider defaults. Summary-display options affect the recorded
trace and do not specify a reasoning-token budget. Gemini uses streaming JSON
output; its TBLite token ledgers identify \texttt{gemini-3.5-flash} as the served
model and \texttt{gemini-3-flash-preview} as a helper, although the original
request used the alias \texttt{gemini-3.8-flash}.

\subsection{Downstream Evaluation and Result Aggregation}
\label{app:eval}\label{app:terminal-outcome-audit}\label{app:analysis}
For TBLite, the model proposes terminal commands at each step; Terminus-2 executes
them and returns the terminal output for the model's next step. Harbor
manages the task environment and runs the verifier after the attempt.
The conversation retains the model's reasoning between command executions. A vLLM server
runs the evaluated checkpoint. The configured temperature is 0.6, top-$p$ is 0.95,
and top-$k$ is 20. The harness model limits are 32,768 input tokens,
40,960 total tokens, and 7,168 output tokens. These are model-call limits,
not a total token budget for an entire task. The task-specific agent time limits range from 300 to 3,600 seconds;
the agent timeout multiplier is one. Deployments use tensor parallelism
2.

BFCL evaluation uses a 65,536-token context and temperature 0.7.
Each of the three evaluation seeds covers the full 800-case BFCL V3
multi-turn suite.

We average evaluation scores within
each checkpoint before computing run-level metrics, then average these
metrics across independent selection runs. Group-performance standard
deviations measure variation across selection runs. TBLite checkpoint
scores include all 300 task attempts. Ranking
metrics compare all five groups within a valid selection run. Failed
submissions receive the fallback scores defined in
Section~\ref{sec:evaluation-metrics} and remain in the run-level averages. Random selection
uses one fixed subset per task; its repeated evaluations measure
variability for that subset.
Full selection-level results appear in Appendix~\ref{app:detailed-results}.

\subsection{Selection-Time Checks and Recipe Information}
\label{app:raw-audit}
A Python guard disables common training entry points, including
\texttt{Tensor.backward} and \texttt{Trainer.train}, while allowing forward
passes. Model-file metadata and training-related trace terms support
further review. Metadata are not weight-content hashes, and keyword
matches can arise from reading recipes; these checks alone do not certify
that every access or training restriction was enforced.
We retain available CLI/tool traces, analysis scripts, intermediate
summaries, final assignments, method reports, and validation outputs.
Timestamps, commands, outputs, and exit status establish the execution
sequence against which agent-written claims are checked.
Appendix~\ref{app:accepted-methods} summarizes selection procedures and
differences in record coverage.

Within each task, agents receive the same fixed training recipe, and all
selected groups are fine-tuned using that recipe. The training procedure
does not vary across agents or selection runs.

\subsection{Cost and Runtime Measurement}
\label{app:runtime}
Selection costs in Table~\ref{tab:api-cost} are averaged across three runs
per agent and task. The BFCL entries use the recorded model-token ledgers
and invocation durations for the 18 selections. Multiple recorded
invocations belonging to the same selection are combined before averaging
the three runs. Durations include tool execution and waits within those
invocations, but exclude gaps between invocations and scheduler queueing.
Astra lacks complete duration and usage coverage; its cost is therefore
reported as a lower bound and its duration is omitted. Token estimates
use the same fixed rate schedule as the TBLite comparison. GPU inference, SFT, and evaluation
costs are excluded. DeepSeek ranges vary peak/off-peak rates; Gemini ranges
vary the billing treatment of unclassified output tokens.
For reproducibility, the fixed USD-per-million-token rates for fresh input,
cache reads, cache writes, and output are $(10,0.25,12.5,50)$ for Fable 5.1,
$(5,0.5,6.25,25)$ for Opus 5, $(10,1,12.5,50)$ for Astra, and
$(3,0.3,3,15)$ for Kimi K3. One-hour Claude cache writes use twice the
fresh-input rate. DeepSeek uses $(0.66,0.022,0,1.98)$ to twice those rates.
Gemini Flash uses $(1.5,0.15,0,9)$, and its separately logged helper model
uses $(0.5,0.05,0,3)$. These are comparison assumptions, not measured charges.

Training times in Table~\ref{tab:api-cost} exclude queueing and model
loading. The TBLite sample comprises 16 completed four-GPU jobs, with a
median of 2.04 h and a range of 1.38--3.30 h. The BFCL sample comprises
90 completed 50-trajectory runs, with a median of 0.29 h and an
interquartile range of 0.25--0.38 h. BFCL evaluation timing covers 35 completed single-GPU jobs, each evaluating
one checkpoint across all three evaluation seeds through a shared model
server. The median is 3.30 h, including model startup and excluding queueing.

\section{Detailed Experimental Results}
\label{app:experimental-results}
\subsection{Selection Procedures}
\label{app:accepted-methods}
The TBLite summaries below distinguish signals used in the final method from the
rule that allocates groups. Runs 1--3 identify each agent's three reported selection attempts in order;
they are not training seeds. NLL denotes base-model negative log-likelihood
on assistant tokens. Completion and verification features are proxies
extracted from trajectories, not access to downstream evaluation outcomes.
``Consecutive blocks'' means partitioning the selected score order into
five equal groups; ``bands'' means selecting separated portions of a ranking.
Each row describes one selection attempt, with its signals and final grouping rule.
Full execution records and scripts are retained separately from these summaries.

Figure~\ref{fig:case-traces} illustrates one Terminal procedure in detail.
Each group contains 1,000 trajectories. The descriptions below summarize the executed selection code.
{\small\setlength{\tabcolsep}{4pt}\renewcommand{\arraystretch}{1.06}
\begin{longtable}{@{}>{\raggedright\arraybackslash}p{.13\linewidth}>{\raggedright\arraybackslash}p{.04\linewidth}>{\raggedright\arraybackslash}p{.36\linewidth}>{\raggedright\arraybackslash}p{.39\linewidth}@{}}
\caption{TBLite selection methods, with one row per attempt.}\label{tab:terminal-method-summary}\\
\toprule Agent & Run & Signals used & Final grouping rule \\\midrule\endfirsthead
\multicolumn{4}{l}{\tablename\ \thetable\ continued}\\
\toprule Agent & Run & Signals used & Final grouping rule \\\midrule\endhead
\midrule\multicolumn{4}{r}{Continued on next page}\\\endfoot
\bottomrule\endlastfoot
\multirow[t]{3}{=}{\textbf{Fable 5}} & 1 & Structure, NLL tails and 64 task clusters. & Cluster-balanced \ensuremath{g_1}; spaced lower score bands. \\*[3pt]
 & 2 & Structure and partial NLL coverage. & Task deduplication and cluster caps in \ensuremath{g_1}. \\*[3pt]
 & 3 & Structure and 8K-prefix NLL. & Task-deduplicated \ensuremath{g_1} without clustering; lower score windows. \\[3pt]
\midrule
\multirow[t]{3}{=}{\textbf{Fable 5.1}} & 1 & Structural quality with a small JSON-NLL term. & Task deduplication and cluster/style caps in \ensuremath{g_1}/\ensuremath{g_2}; lower score bands. \\*[3pt]
 & 2 & Structural quality, model loss and a prompted concreteness score. & Task deduplication and cluster/style caps in \ensuremath{g_1}/\ensuremath{g_2}; lower score bands. \\*[3pt]
 & 3 & Structural quality with soft loss-anomaly penalties. & Cluster/source caps and an 8\% Tezos cap in \ensuremath{g_1}/\ensuremath{g_2}; next-ranked \ensuremath{g_3}/\ensuremath{g_4} and bottom \ensuremath{g_5}. \\[3pt]
\midrule
\multirow[t]{3}{=}{\textbf{Opus 5}} & 1 & Completion, protocol and verification features plus model statistics. & \ensuremath{g_1} eligibility checks, task deduplication and cluster caps; lower score bands. \\*[3pt]
 & 2 & Quality, NLL and embeddings. & Diversity-aware greedy \ensuremath{g_1}; lower score bands. \\*[3pt]
 & 3 & Execution evidence and length-adjusted action loss. & \ensuremath{g_1} eligibility with soft similarity, cluster and source penalties; lower score bands. \\[3pt]
\midrule
\multirow[t]{3}{=}{\textbf{Sonnet 5}} & 1 & Completion, format and 4K-prefix loss. & Top 5,000 split into consecutive groups; no hard task deduplication. \\*[3pt]
 & 2 & Structural quality and preference for median loss. & Retains one example per task hash, then splits the top 5,000 into consecutive groups. \\*[3pt]
 & 3 & Format, loss and turn efficiency. & Heap follows global score order; top 5,000 split consecutively without hard task deduplication. \\[3pt]
\midrule
\multirow[t]{3}{=}{\textbf{GPT-5.5}} & 1 & Structure, candidate NLL and task-term rarity. & Top 5,000 split consecutively; soft novelty adjustments without hard task deduplication. \\*[3pt]
 & 2 & Structure and 2K-prefix NLL; domain counts do not affect selection. & Top 5,000 split consecutively; no hard task deduplication. \\*[3pt]
 & 3 & Structural scoring, then NLL reranking of the top 3,000. & Top 5,000 split consecutively; no hard task deduplication. \\[3pt]
\midrule
\multirow[t]{3}{=}{\textbf{GPT-5.6 Sol}} & 1 & Completion, action, test/build and supervision features; tokenized lengths, no model loss. & Top 5,000 split consecutively without hard task deduplication. \\*[3pt]
 & 2 & Completion, action, test/build and supervision features; estimated truncation, no model loss. & Top 5,000 split consecutively without hard task deduplication. \\*[3pt]
 & 3 & Completion, action, test/build and supervision features; tokenized lengths, no model loss. & Top 5,000 split consecutively; ties use test/coding evidence, length and row index. No hard task deduplication. \\[3pt]
\midrule
\multirow[t]{3}{=}{\textbf{GPT-6 Astra}} & 1 & Prompted quality judgments, execution evidence, sampled NLL and task representations; staged review. & Diverse \ensuremath{g_1} with task deduplication and soft redundancy/coverage penalties; separated lower bands. \\*[3pt]
 & 2 & Prompted quality judgments, execution evidence, sampled NLL and task representations; three assessments. & Diverse \ensuremath{g_1} with task deduplication and soft redundancy/coverage penalties; separated lower bands. \\*[3pt]
 & 3 & Prompted quality judgments, execution evidence, sampled NLL and task representations; five-level rubric. & \ensuremath{g_1} completion/protocol/context eligibility, task deduplication and soft redundancy/coverage penalties; lower bands. \\[3pt]
\midrule
\multirow[t]{3}{=}{\textbf{DeepSeek V4 Pro}} & 1 & Structural value and embeddings; NLL discarded. & Similarity-penalized greedy order. \\*[3pt]
 & 2 & Loss weighted by cleanliness and completion. & One example per task hash, then top-5,000 consecutive blocks. \\*[3pt]
 & 3 & Structural activity and recovery; no NLL in the final score. & Completed candidates, TF-IDF similarity filter at 0.92, then consecutive blocks. \\[3pt]
\midrule
\multirow[t]{3}{=}{\textbf{Kimi K3}} & 1 & Completion, supervision and a small NLL term. & Task deduplication and \ensuremath{g_1} cluster rebalancing. \\*[3pt]
 & 2 & Preference for moderate length-residual loss. & Similarity threshold and up to three examples per task in \ensuremath{g_1}. \\*[3pt]
 & 3 & Negative length-normalized NLL plus completion/truncation terms. & Cosine filter at 0.97 with fallback, then score-sorted blocks. \\[3pt]
\midrule
\multirow[t]{3}{=}{\textbf{Gemini 3.5 Flash}} & 1 & Multiplicative structural score. & Top 5,000 completed examples, split into consecutive blocks. \\*[3pt]
 & 2 & Additive completion/error/length score. & Top 4,000 for \ensuremath{g_1}--\ensuremath{g_4}; bottom 1,000 for \ensuremath{g_5}. \\*[3pt]
 & 3 & NLL and thinking density on a premium candidate pool. & Premium \ensuremath{g_1}/\ensuremath{g_2}, other completed \ensuremath{g_3}, and stronger/weaker incomplete \ensuremath{g_4}/\ensuremath{g_5}. \\[3pt]
\end{longtable}}

\subsection{BFCL Selection Procedures}
We inspect the final method reports together with selection scripts or
command traces for the 18 runs using 50-trajectory groups. The descriptions
below summarize the rules used to form the submitted groups. Quality,
correctness, and difficulty refer to the agents' selection signals,
not independent labels of training utility.

\begin{table}[!htbp]
\centering\small
\setlength{\tabcolsep}{4pt}
\renewcommand{\arraystretch}{1.08}
\caption{BFCL selection recipes and differences across the three independent
runs. Each run produces five disjoint groups of 50 complete trajectories.}
\label{tab:bfcl-method-summary}
\begin{tabular}{@{}p{.14\linewidth}p{.81\linewidth}@{}}\toprule
Agent & Recipe and cross-run differences\\\midrule
Fable 5.1 &
Run 1 combines schema checks, errors, reasoning length, and length-adjusted
loss; it samples descending score bands, ending with low-scoring rows.
Run 2 emphasizes reasoning--action consistency and loss outliers.
Run 3 separately scores reasoning, actions given reasoning, and actions
without reasoning, then combines quality gates with environment coverage.
Later groups include increasingly defect-heavy trajectories.\\[3pt]
Opus 5 &
Run 1 prioritizes structural quality and uses a small residualized-loss term,
with fixed interaction-type and environment coverage across groups.
Run 2 adds verbosity-normalized reasoning loss and recovered-error bonuses,
placing noncanonical argument formats in the lower groups.
Run 3 probes action difficulty without the demonstrated reasoning and
selects widely separated score bands with distinct environments.\\[3pt]
GPT-6 Astra &
All runs combine schema checks, argument grounding, state-changing actions,
clarification, and coverage. Run 1 adds sampled-turn loss over the pool and
inspection of top candidates. Run 2 uses a bounded preference for moderate
reasoning loss and a fixed 70/30 interaction-type mix. Run 3 scores a
quality-filtered shortlist and adds semantic checks on unsupported actions.
Groups occupy separated predicted-value bands.\\[3pt]
Kimi K3 &
Run 1 uses loss to penalize outliers, normalizes quality within interaction
types, and selects successive groups under diversity caps. Run 2 favors
intermediate total and tool-call loss. Run 3 instead rewards lower loss
alongside successful responses and tool coverage, with environment caps
relaxed from one to three across the groups.\\[3pt]
DeepSeek V4 Pro &
Run 1 prioritizes high sampled-turn loss, trajectory length, and final-action
loss. Run 2 combines loss with structural richness, including a positive
failed-call term, and uses diversity-penalized greedy selection. Run 3
emphasizes tool breadth, penalizes failed-call rate, and adds
within-environment loss as a secondary signal.\\[3pt]
Gemini 3.5 Flash &
Run 1 separates groups by observed tool success, length, and turn count.
Run 2 uses a heuristic score penalizing errors and loops while rewarding
reasoning length and intermediate turn counts. Run 3 adds base-model loss
to choose long, error-free trajectories for the top groups; short clean
trajectories and error-heavy trajectories occupy lower groups.\\
\bottomrule\end{tabular}
\end{table}

\subsection{Results by Selection}
\label{app:detailed-results}
These tables retain the individual selections underlying the averages in
the main text. Scores across evaluation seeds are averaged within each checkpoint;
ranking metrics are computed within each selection before aggregation.
\begin{table}[!htbp]
\centering\normalsize
\setlength{\tabcolsep}{4pt}\setlength{\fboxsep}{1.4pt}
\caption{Terminal group-level results on a 0--1 reward scale. Runs use the same 1--3 numbering as the procedure summaries. Each reward averages 100 tasks with three attempts per task. Within each row, shading runs from darkest (best) to lightest (worst), and bold marks the best group. Colors use unrounded means; equal means receive the same shade. $\rho$ compares the predicted \ensuremath{g_1}--\ensuremath{g_5} order with observed performance; \ensuremath{g_1} rank is one plus the number of groups scoring strictly higher.}
\label{tab:terminal-orders}
\begin{tabular}{@{}llrrrrrrr@{}}\toprule
Agent & Run & \ensuremath{g_1} & \ensuremath{g_2} & \ensuremath{g_3} & \ensuremath{g_4} & \ensuremath{g_5} & $\rho$ & \ensuremath{g_1} rank \\\midrule
Fable 5 & r1 & \colorbox{blue!22}{\makebox[25pt][r]{0.109}} & \colorbox{blue!4}{\makebox[25pt][r]{0.085}} & \colorbox{blue!28}{\makebox[25pt][r]{\textbf{0.145}}} & \colorbox{blue!10}{\makebox[25pt][r]{0.101}} & \colorbox{blue!16}{\makebox[25pt][r]{0.107}} & 0.10 & 2 \\
 & r2 & \colorbox{blue!10}{\makebox[25pt][r]{0.100}} & \colorbox{blue!22}{\makebox[25pt][r]{0.110}} & \colorbox{blue!28}{\makebox[25pt][r]{\textbf{0.121}}} & \colorbox{blue!16}{\makebox[25pt][r]{0.110}} & \colorbox{blue!4}{\makebox[25pt][r]{0.099}} & 0.30 & 4 \\
 & r3 & \colorbox{blue!10}{\makebox[25pt][r]{0.121}} & \colorbox{blue!4}{\makebox[25pt][r]{0.121}} & \colorbox{blue!28}{\makebox[25pt][r]{\textbf{0.135}}} & \colorbox{blue!22}{\makebox[25pt][r]{0.134}} & \colorbox{blue!16}{\makebox[25pt][r]{0.121}} & -0.50 & 4 \\
\addlinespace[3pt]
Fable 5.1 & r1 & \colorbox{blue!16}{\makebox[25pt][r]{0.117}} & \colorbox{blue!22}{\makebox[25pt][r]{0.122}} & \colorbox{blue!28}{\makebox[25pt][r]{\textbf{0.129}}} & \colorbox{blue!4}{\makebox[25pt][r]{0.098}} & \colorbox{blue!10}{\makebox[25pt][r]{0.116}} & 0.50 & 3 \\
 & r2 & \colorbox{blue!16}{\makebox[25pt][r]{0.110}} & \colorbox{blue!10}{\makebox[25pt][r]{0.103}} & \colorbox{blue!22}{\makebox[25pt][r]{0.110}} & \colorbox{blue!4}{\makebox[25pt][r]{0.097}} & \colorbox{blue!28}{\makebox[25pt][r]{\textbf{0.132}}} & -0.30 & 3 \\
 & r3 & \colorbox{blue!28}{\makebox[25pt][r]{\textbf{0.148}}} & \colorbox{blue!10}{\makebox[25pt][r]{0.111}} & \colorbox{blue!22}{\makebox[25pt][r]{0.121}} & \colorbox{blue!16}{\makebox[25pt][r]{0.112}} & \colorbox{blue!4}{\makebox[25pt][r]{0.101}} & 0.70 & 1 \\
\addlinespace[3pt]
Opus 5 & r1 & \colorbox{blue!16}{\makebox[25pt][r]{0.125}} & \colorbox{blue!28}{\makebox[25pt][r]{\textbf{0.137}}} & \colorbox{blue!22}{\makebox[25pt][r]{0.137}} & \colorbox{blue!4}{\makebox[25pt][r]{0.094}} & \colorbox{blue!10}{\makebox[25pt][r]{0.113}} & 0.60 & 3 \\
 & r2 & \colorbox{blue!10}{\makebox[25pt][r]{0.115}} & \colorbox{blue!22}{\makebox[25pt][r]{0.127}} & \colorbox{blue!4}{\makebox[25pt][r]{0.097}} & \colorbox{blue!28}{\makebox[25pt][r]{\textbf{0.135}}} & \colorbox{blue!16}{\makebox[25pt][r]{0.117}} & -0.30 & 4 \\
 & r3 & \colorbox{blue!16}{\makebox[25pt][r]{0.108}} & \colorbox{blue!22}{\makebox[25pt][r]{0.111}} & \colorbox{blue!28}{\makebox[25pt][r]{\textbf{0.124}}} & \colorbox{blue!10}{\makebox[25pt][r]{0.099}} & \colorbox{blue!4}{\makebox[25pt][r]{0.081}} & 0.60 & 3 \\
\addlinespace[3pt]
Sonnet 5 & r1 & \colorbox{blue!4}{\makebox[25pt][r]{0.099}} & \colorbox{blue!10}{\makebox[25pt][r]{0.110}} & \colorbox{blue!16}{\makebox[25pt][r]{0.116}} & \colorbox{blue!22}{\makebox[25pt][r]{0.124}} & \colorbox{blue!28}{\makebox[25pt][r]{\textbf{0.149}}} & -1.00 & 5 \\
 & r2 & \colorbox{blue!16}{\makebox[25pt][r]{0.118}} & \colorbox{blue!22}{\makebox[25pt][r]{0.123}} & \colorbox{blue!4}{\makebox[25pt][r]{0.101}} & \colorbox{blue!10}{\makebox[25pt][r]{0.117}} & \colorbox{blue!28}{\makebox[25pt][r]{\textbf{0.124}}} & -0.20 & 3 \\
 & r3 & \colorbox{blue!28}{\makebox[25pt][r]{\textbf{0.130}}} & \colorbox{blue!22}{\makebox[25pt][r]{0.125}} & \colorbox{blue!4}{\makebox[25pt][r]{0.103}} & \colorbox{blue!10}{\makebox[25pt][r]{0.113}} & \colorbox{blue!16}{\makebox[25pt][r]{0.115}} & 0.60 & 1 \\
\addlinespace[3pt]
GPT-5.5 & r1 & \colorbox{blue!16}{\makebox[25pt][r]{0.107}} & \colorbox{blue!4}{\makebox[25pt][r]{0.086}} & \colorbox{blue!22}{\makebox[25pt][r]{0.109}} & \colorbox{blue!10}{\makebox[25pt][r]{0.099}} & \colorbox{blue!28}{\makebox[25pt][r]{\textbf{0.126}}} & -0.50 & 3 \\
 & r2 & \colorbox{blue!22}{\makebox[25pt][r]{0.128}} & \colorbox{blue!4}{\makebox[25pt][r]{0.103}} & \colorbox{blue!28}{\makebox[25pt][r]{\textbf{0.133}}} & \colorbox{blue!16}{\makebox[25pt][r]{0.127}} & \colorbox{blue!10}{\makebox[25pt][r]{0.122}} & 0.20 & 2 \\
 & r3 & \colorbox{blue!22}{\makebox[25pt][r]{0.120}} & \colorbox{blue!4}{\makebox[25pt][r]{0.106}} & \colorbox{blue!10}{\makebox[25pt][r]{0.109}} & \colorbox{blue!16}{\makebox[25pt][r]{0.120}} & \colorbox{blue!28}{\makebox[25pt][r]{\textbf{0.125}}} & -0.40 & 2 \\
\addlinespace[3pt]
GPT-5.6 Sol & r1 & \colorbox{blue!16}{\makebox[25pt][r]{0.130}} & \colorbox{blue!4}{\makebox[25pt][r]{0.111}} & \colorbox{blue!10}{\makebox[25pt][r]{0.126}} & \colorbox{blue!22}{\makebox[25pt][r]{0.133}} & \colorbox{blue!28}{\makebox[25pt][r]{\textbf{0.148}}} & -0.70 & 3 \\
 & r2 & \colorbox{blue!10}{\makebox[25pt][r]{0.101}} & \colorbox{blue!22}{\makebox[25pt][r]{0.108}} & \colorbox{blue!16}{\makebox[25pt][r]{0.105}} & \colorbox{blue!28}{\makebox[25pt][r]{\textbf{0.123}}} & \colorbox{blue!4}{\makebox[25pt][r]{0.100}} & 0.10 & 4 \\
 & r3 & \colorbox{blue!4}{\makebox[25pt][r]{0.107}} & \colorbox{blue!16}{\makebox[25pt][r]{0.118}} & \colorbox{blue!10}{\makebox[25pt][r]{0.111}} & \colorbox{blue!22}{\makebox[25pt][r]{0.131}} & \colorbox{blue!28}{\makebox[25pt][r]{\textbf{0.135}}} & -0.90 & 5 \\
\addlinespace[3pt]
GPT-6 Astra & r1 & \colorbox{blue!4}{\makebox[25pt][r]{0.112}} & \colorbox{blue!28}{\makebox[25pt][r]{\textbf{0.143}}} & \colorbox{blue!10}{\makebox[25pt][r]{0.113}} & \colorbox{blue!16}{\makebox[25pt][r]{0.114}} & \colorbox{blue!22}{\makebox[25pt][r]{0.121}} & -0.40 & 5 \\
 & r2 & \colorbox{blue!10}{\makebox[25pt][r]{0.105}} & \colorbox{blue!28}{\makebox[25pt][r]{\textbf{0.145}}} & \colorbox{blue!4}{\makebox[25pt][r]{0.097}} & \colorbox{blue!22}{\makebox[25pt][r]{0.136}} & \colorbox{blue!16}{\makebox[25pt][r]{0.114}} & -0.10 & 4 \\
 & r3 & \colorbox{blue!28}{\makebox[25pt][r]{\textbf{0.142}}} & \colorbox{blue!22}{\makebox[25pt][r]{0.112}} & \colorbox{blue!16}{\makebox[25pt][r]{0.108}} & \colorbox{blue!10}{\makebox[25pt][r]{0.098}} & \colorbox{blue!4}{\makebox[25pt][r]{0.078}} & 1.00 & 1 \\
\addlinespace[3pt]
DeepSeek V4 Pro & r1 & \colorbox{blue!28}{\makebox[25pt][r]{\textbf{0.129}}} & \colorbox{blue!4}{\makebox[25pt][r]{0.098}} & \colorbox{blue!16}{\makebox[25pt][r]{0.109}} & \colorbox{blue!22}{\makebox[25pt][r]{0.125}} & \colorbox{blue!10}{\makebox[25pt][r]{0.099}} & 0.30 & 1 \\
 & r2 & \colorbox{blue!28}{\makebox[25pt][r]{\textbf{0.135}}} & \colorbox{blue!22}{\makebox[25pt][r]{0.133}} & \colorbox{blue!4}{\makebox[25pt][r]{0.104}} & \colorbox{blue!10}{\makebox[25pt][r]{0.120}} & \colorbox{blue!16}{\makebox[25pt][r]{0.132}} & 0.60 & 1 \\
 & r3 & \colorbox{blue!4}{\makebox[25pt][r]{0.112}} & \colorbox{blue!16}{\makebox[25pt][r]{0.122}} & \colorbox{blue!28}{\makebox[25pt][r]{\textbf{0.137}}} & \colorbox{blue!10}{\makebox[25pt][r]{0.121}} & \colorbox{blue!22}{\makebox[25pt][r]{0.132}} & -0.50 & 5 \\
\addlinespace[3pt]
Kimi K3 & r1 & \colorbox{blue!28}{\makebox[25pt][r]{\textbf{0.134}}} & \colorbox{blue!22}{\makebox[25pt][r]{0.131}} & \colorbox{blue!4}{\makebox[25pt][r]{0.112}} & \colorbox{blue!16}{\makebox[25pt][r]{0.130}} & \colorbox{blue!10}{\makebox[25pt][r]{0.114}} & 0.70 & 1 \\
 & r2 & \colorbox{blue!10}{\makebox[25pt][r]{0.105}} & \colorbox{blue!4}{\makebox[25pt][r]{0.091}} & \colorbox{blue!22}{\makebox[25pt][r]{0.108}} & \colorbox{blue!28}{\makebox[25pt][r]{\textbf{0.110}}} & \colorbox{blue!16}{\makebox[25pt][r]{0.107}} & -0.60 & 4 \\
 & r3 & \colorbox{blue!16}{\makebox[25pt][r]{0.119}} & \colorbox{blue!10}{\makebox[25pt][r]{0.109}} & \colorbox{blue!22}{\makebox[25pt][r]{0.120}} & \colorbox{blue!28}{\makebox[25pt][r]{\textbf{0.152}}} & \colorbox{blue!4}{\makebox[25pt][r]{0.103}} & 0.10 & 3 \\
\addlinespace[3pt]
Gemini 3.5 Flash & r1 & \colorbox{blue!10}{\makebox[25pt][r]{0.115}} & \colorbox{blue!22}{\makebox[25pt][r]{0.137}} & \colorbox{blue!4}{\makebox[25pt][r]{0.102}} & \colorbox{blue!28}{\makebox[25pt][r]{\textbf{0.137}}} & \colorbox{blue!16}{\makebox[25pt][r]{0.135}} & -0.30 & 4 \\
 & r2 & \colorbox{blue!10}{\makebox[25pt][r]{0.117}} & \colorbox{blue!22}{\makebox[25pt][r]{0.127}} & \colorbox{blue!28}{\makebox[25pt][r]{\textbf{0.132}}} & \colorbox{blue!16}{\makebox[25pt][r]{0.121}} & \colorbox{blue!4}{\makebox[25pt][r]{0.112}} & 0.30 & 4 \\
 & r3 & \colorbox{blue!22}{\makebox[25pt][r]{0.130}} & \colorbox{blue!28}{\makebox[25pt][r]{\textbf{0.139}}} & \colorbox{blue!16}{\makebox[25pt][r]{0.120}} & \colorbox{blue!10}{\makebox[25pt][r]{0.119}} & \colorbox{blue!4}{\makebox[25pt][r]{0.082}} & 0.90 & 2 \\
\addlinespace[3pt]
\bottomrule\end{tabular}\end{table}

\begin{table}[!htbp]
\centering\small
\caption{BFCL accuracy (\%) for 50-trajectory groups in each independent selection run.
Scores average the 800-case evaluations across evaluation seeds. Shading compares groups within a run;
$\rho$ measures full-order agreement; $g_1$ rank is one plus the number of groups scoring strictly higher.}
\label{tab:bfcl-orders}
\setlength{\tabcolsep}{3pt}
\begin{tabular}{@{}llrrrrrrr@{}}\toprule
Agent & Run & $g_1$ & $g_2$ & $g_3$ & $g_4$ & $g_5$ & $\rho$ & $g_1$ rank\\\midrule
Fable 5.1 & r1 & \colorbox{blue!22}{\makebox[25pt][r]{36.00}} & \colorbox{blue!4}{\makebox[25pt][r]{33.25}} & \colorbox{blue!16}{\makebox[25pt][r]{34.83}} & \colorbox{blue!10}{\makebox[25pt][r]{34.38}} & \colorbox{blue!28}{\makebox[25pt][r]{\textbf{37.17}}} & -0.30 & 2 \\
 & r2 & \colorbox{blue!16}{\makebox[25pt][r]{34.17}} & \colorbox{blue!28}{\makebox[25pt][r]{\textbf{36.58}}} & \colorbox{blue!22}{\makebox[25pt][r]{34.96}} & \colorbox{blue!4}{\makebox[25pt][r]{33.96}} & \colorbox{blue!10}{\makebox[25pt][r]{34.12}} & 0.60 & 3 \\
 & r3 & \colorbox{blue!28}{\makebox[25pt][r]{\textbf{35.46}}} & \colorbox{blue!16}{\makebox[25pt][r]{35.04}} & \colorbox{blue!10}{\makebox[25pt][r]{34.83}} & \colorbox{blue!4}{\makebox[25pt][r]{33.96}} & \colorbox{blue!22}{\makebox[25pt][r]{35.42}} & 0.40 & 1 \\
\addlinespace[3pt]
Opus 5 & r1 & \colorbox{blue!22}{\makebox[25pt][r]{35.71}} & \colorbox{blue!28}{\makebox[25pt][r]{\textbf{36.46}}} & \colorbox{blue!16}{\makebox[25pt][r]{35.08}} & \colorbox{blue!10}{\makebox[25pt][r]{31.96}} & \colorbox{blue!4}{\makebox[25pt][r]{28.75}} & 0.90 & 2 \\
 & r2 & \colorbox{blue!22}{\makebox[25pt][r]{36.75}} & \colorbox{blue!28}{\makebox[25pt][r]{\textbf{37.38}}} & \colorbox{blue!10}{\makebox[25pt][r]{35.00}} & \colorbox{blue!16}{\makebox[25pt][r]{36.54}} & \colorbox{blue!4}{\makebox[25pt][r]{29.46}} & 0.80 & 2 \\
 & r3 & \colorbox{blue!22}{\makebox[25pt][r]{35.38}} & \colorbox{blue!16}{\makebox[25pt][r]{35.21}} & \colorbox{blue!28}{\makebox[25pt][r]{\textbf{35.96}}} & \colorbox{blue!4}{\makebox[25pt][r]{34.38}} & \colorbox{blue!10}{\makebox[25pt][r]{35.04}} & 0.60 & 2 \\
\addlinespace[3pt]
GPT-6 Astra & r1 & \colorbox{blue!28}{\makebox[25pt][r]{\textbf{36.38}}} & \colorbox{blue!16}{\makebox[25pt][r]{34.58}} & \colorbox{blue!4}{\makebox[25pt][r]{33.21}} & \colorbox{blue!22}{\makebox[25pt][r]{34.71}} & \colorbox{blue!10}{\makebox[25pt][r]{34.50}} & 0.50 & 1 \\
 & r2 & \colorbox{blue!28}{\makebox[25pt][r]{\textbf{36.75}}} & \colorbox{blue!22}{\makebox[25pt][r]{36.38}} & \colorbox{blue!16}{\makebox[25pt][r]{36.21}} & \colorbox{blue!10}{\makebox[25pt][r]{35.42}} & \colorbox{blue!4}{\makebox[25pt][r]{34.62}} & 1.00 & 1 \\
 & r3 & \colorbox{blue!28}{\makebox[25pt][r]{\textbf{36.33}}} & \colorbox{blue!22}{\makebox[25pt][r]{34.96}} & \colorbox{blue!16}{\makebox[25pt][r]{34.79}} & \colorbox{blue!10}{\makebox[25pt][r]{34.67}} & \colorbox{blue!4}{\makebox[25pt][r]{34.62}} & 1.00 & 1 \\
\addlinespace[3pt]
DeepSeek V4 Pro & r1 & \colorbox{blue!13}{\makebox[25pt][r]{35.29}} & \colorbox{blue!28}{\makebox[25pt][r]{\textbf{37.42}}} & \colorbox{blue!22}{\makebox[25pt][r]{35.33}} & \colorbox{blue!13}{\makebox[25pt][r]{35.29}} & \colorbox{blue!4}{\makebox[25pt][r]{33.58}} & 0.56 & 3 \\
 & r2 & \colorbox{blue!22}{\makebox[25pt][r]{35.67}} & \colorbox{blue!28}{\makebox[25pt][r]{\textbf{36.04}}} & \colorbox{blue!10}{\makebox[25pt][r]{34.88}} & \colorbox{blue!16}{\makebox[25pt][r]{35.21}} & \colorbox{blue!4}{\makebox[25pt][r]{33.75}} & 0.80 & 2 \\
 & r3 & \colorbox{blue!28}{\makebox[25pt][r]{\textbf{38.62}}} & \colorbox{blue!10}{\makebox[25pt][r]{36.38}} & \colorbox{blue!16}{\makebox[25pt][r]{37.00}} & \colorbox{blue!22}{\makebox[25pt][r]{37.75}} & \colorbox{blue!4}{\makebox[25pt][r]{35.67}} & 0.60 & 1 \\
\addlinespace[3pt]
Kimi K3 & r1 & \colorbox{blue!10}{\makebox[25pt][r]{35.25}} & \colorbox{blue!22}{\makebox[25pt][r]{36.08}} & \colorbox{blue!16}{\makebox[25pt][r]{35.42}} & \colorbox{blue!4}{\makebox[25pt][r]{34.54}} & \colorbox{blue!28}{\makebox[25pt][r]{\textbf{36.42}}} & -0.30 & 4 \\
 & r2 & \colorbox{blue!16}{\makebox[25pt][r]{36.71}} & \colorbox{blue!4}{\makebox[25pt][r]{35.62}} & \colorbox{blue!22}{\makebox[25pt][r]{37.58}} & \colorbox{blue!10}{\makebox[25pt][r]{36.38}} & \colorbox{blue!28}{\makebox[25pt][r]{\textbf{37.75}}} & -0.50 & 3 \\
 & r3 & \colorbox{blue!25}{\makebox[25pt][r]{\textbf{36.92}}} & \colorbox{blue!25}{\makebox[25pt][r]{\textbf{36.92}}} & \colorbox{blue!10}{\makebox[25pt][r]{35.54}} & \colorbox{blue!16}{\makebox[25pt][r]{36.62}} & \colorbox{blue!4}{\makebox[25pt][r]{35.21}} & 0.87 & 1 \\
\addlinespace[3pt]
Gemini 3.5 Flash & r1 & \colorbox{blue!4}{\makebox[25pt][r]{33.38}} & \colorbox{blue!16}{\makebox[25pt][r]{34.17}} & \colorbox{blue!10}{\makebox[25pt][r]{34.12}} & \colorbox{blue!28}{\makebox[25pt][r]{\textbf{35.21}}} & \colorbox{blue!22}{\makebox[25pt][r]{34.54}} & -0.80 & 5 \\
 & r2 & \colorbox{blue!22}{\makebox[25pt][r]{35.50}} & \colorbox{blue!4}{\makebox[25pt][r]{33.08}} & \colorbox{blue!10}{\makebox[25pt][r]{33.21}} & \colorbox{blue!16}{\makebox[25pt][r]{34.79}} & \colorbox{blue!28}{\makebox[25pt][r]{\textbf{35.92}}} & -0.40 & 2 \\
 & r3 & \colorbox{blue!28}{\makebox[25pt][r]{\textbf{36.42}}} & \colorbox{blue!16}{\makebox[25pt][r]{35.29}} & \colorbox{blue!10}{\makebox[25pt][r]{33.38}} & \colorbox{blue!4}{\makebox[25pt][r]{29.21}} & \colorbox{blue!22}{\makebox[25pt][r]{36.33}} & 0.40 & 1 \\
\addlinespace[3pt]
\bottomrule\end{tabular}
\end{table}

\FloatBarrier

\clearpage
\section{Selection Prompts}
\label{app:prompt}
\definecolor{promptpaper}{HTML}{FAF8F3}
\definecolor{promptborder}{HTML}{D9B9A4}
\tcbset{promptstyle/.style={enhanced,breakable,colback=promptpaper,
  colframe=promptborder,boxrule=0.4pt,arc=1mm,
  left=6pt,right=6pt,top=5pt,bottom=5pt,before skip=7pt,after skip=7pt}}
\makeatletter
\newenvironment{selectionprompt}{%
  \begin{tcolorbox}[promptstyle]
  \small\raggedright
  \setlength{\parindent}{0pt}\setlength{\parskip}{2pt}
  \setlength{\emergencystretch}{2em}
  \def\@listi{\leftmargin\leftmargini\topsep 2pt\parsep 0pt\itemsep 1pt}
}{\end{tcolorbox}}
\makeatother
\newcommand{\promptheading}[1]{\par\addvspace{4pt}\noindent\textbf{#1}\par\nobreak\vspace{1pt}\nobreak}

\subsection{Selection Configuration Examples}
\label{app:config-examples}
Each prompt reads a local \texttt{config.yaml}. The examples below show
the selection-time settings for both tasks, with dataset paths written
relative to the workspace. The BFCL pool contains 8,962 trajectories after
excluding the 60-trajectory validation split. The \texttt{max\_len} field
specifies the length limit supplied during selection; realized training
settings appear in Appendix~\ref{app:recipe}.

\noindent\begin{minipage}[t]{.48\linewidth}
\textbf{Terminal: \texttt{config.yaml}}
\begin{tcolorbox}[promptstyle,unbreakable]
\begin{Verbatim}[fontsize=\small]
base_model: ./model
dataset_path:
  ./data/train.parquet
n_groups: 5
group_size: 1000
max_len: 32768
\end{Verbatim}
\end{tcolorbox}
\end{minipage}\hfill
\begin{minipage}[t]{.48\linewidth}
\textbf{BFCL: \texttt{config.yaml}}
\begin{tcolorbox}[promptstyle,unbreakable]
\begin{Verbatim}[fontsize=\small]
base_model: Qwen/Qwen3-4B
dataset_path:
  ./data/pool.parquet
n_groups: 5
group_size: 50
max_len: 16384
\end{Verbatim}
\end{tcolorbox}
\end{minipage}

\subsection{Terminal Data Selection}
\label{app:prompt-terminal}
All 30 reported terminal selections use the prompt below and the fixed
training recipe described in Appendix~\ref{app:recipe}.
\nolinenumbers
\begin{selectionprompt}
Read \texttt{config.yaml} in the current directory first. It specifies: \texttt{base\_model} (HF id / path), \texttt{dataset\_path} (the full SFT candidate dataset, parquet), \texttt{n\_groups} (=5), \texttt{group\_size} (=1000), and \texttt{max\_len}. Use exactly these --- do not hardcode other values.

We have that base model and the full SFT candidate dataset. A FIXED training recipe (provided as code at \texttt{recipe/sft.py}, do not modify or run it) will later fine-tune the base model separately on each of \texttt{n\_groups} data groups, with identical hyper-parameters; the only variable across runs is which data subset goes in.

\promptheading{Downstream benchmark (what the trained models will be scored on)}
Each trained model will act as a TERMINAL AGENT: dropped into a fresh Unix environment with a natural-language task (e.g. build/fix code, process data, configure the system), it interacts over many turns --- issuing shell commands and reading their output --- and is scored by whether the task's automated tests pass at the end. You never see the benchmark tasks; you only know the evaluation is of this kind.

\promptheading{Your goal}
This is a SELECTION-AND-RANKING task. Predict, for every sample in the dataset, its expected value for SFT (how much it would improve the downstream benchmark if trained on), and use that prediction to SELECT \texttt{n\_groups} groups of exactly \texttt{group\_size} rows each, ordered so that: after training each group with the fixed recipe, benchmark score goes from high to low as the group index goes from 1 to n\_groups (group 1 best, group n\_groups worst).

You are FREE to design the valuation however you think works best. You may use a SINGLE signal or COMBINE MULTIPLE complementary signals (e.g. loss/perplexity, trajectory quality, task difficulty, efficiency, diversity/coverage, length effects, ...) into a composite score --- your choice. Don't feel obliged to use just one textbook metric; use whatever you believe will best achieve the objective below.

\promptheading{Objective (what you are optimizing)}
\begin{enumerate}
\item \textbf{PRIMARY --- make group 1 as strong as possible.} Group 1 should be the single most valuable \texttt{group\_size}-row subset you can assemble from the whole pool: after training on it alone, its benchmark score should be as HIGH as possible --- ideally beating a random subset of the same size and approaching what training on the whole pool would give. Put real effort into getting the very top group right; that is the headline result.
\item \textbf{ORDERING --- rank the groups.} Score should decrease from group 1 to group n\_groups (Spearman between group index and post-trained score -\textgreater{} -1). Place the most valuable samples in the lowest-numbered groups and the least valuable of your selection in the highest.
\end{enumerate}
\promptheading{Deliverable (the ONLY thing you produce)}
Write \texttt{outputs/group\_assignment.csv} with columns: \texttt{index,group\_id}

\begin{itemize}
\item \texttt{index}  = the row index into the dataset (config.dataset\_path) (0-based)
\item \texttt{group\_id} = an integer in 1..n\_groups
\item The file contains ONLY the selected rows: exactly n\_groups x group\_size lines (plus header), each selected index exactly once, exactly \texttt{group\_size} rows per group\_id. Unselected rows must NOT appear in the file.
\end{itemize}
Also write \texttt{outputs/method.md} briefly explaining your method and why. Do NOT produce a trained model. Do NOT run the training recipe.

\promptheading{Hard constraints}
\begin{enumerate}
\item INFERENCE-ONLY. You may only run forward passes of the base model (hidden states / logits are accessible). NO gradients, NO optimizer steps, NO training of any model (not even a small proxy). Gradient/optimizer calls are disabled in this sandbox and will raise.
\item The base model (config.base\_model) is READ-ONLY. Use it as-is; do not fine-tune or overwrite it.
\item Do NOT access any benchmark/test set. Only use the dataset at \texttt{config.dataset\_path}.
\item The n\_groups groups must be mutually disjoint and each contain exactly \texttt{group\_size} rows. They need NOT cover the dataset: rows outside your selection are left unused.
\item Work only in the current directory. Internet is available for packages if needed (transformers/datasets/torch/pyarrow are pre-installed).
\end{enumerate}
\promptheading{Data format}
The dataset at \texttt{config.dataset\_path} has a \texttt{messages} column: a Python list of role/content dicts in chat format; each row is one multi-turn terminal-agent trajectory. The recipe trains on this \texttt{messages} column via the model chat template (tokenizer.apply\_chat\_template), truncating at \texttt{max\_len}, with the non-assistant parts masked so loss is on assistant tokens only. To match it, build inputs with the chat template rather than concatenating raw strings.

\promptheading{Environment}
\begin{itemize}
\item A single H100 GPU (80 GB) is available. The dataset (config.dataset\_path), the base model (config.base\_model), and \texttt{recipe/sft.py} (read-only reference) are in the current directory.
\item TIME BUDGET: complete the task within 4 hours. Plan your computation to fit --- the deliverable must be written and verified before the budget runs out.
\item You operate fully autonomously; never ask for feedback. Just produce \texttt{outputs/group\_assignment.csv}.
\end{itemize}
\promptheading{Execution model (IMPORTANT --- single headless turn)}
You run as ONE non-interactive, single-shot invocation. There is NO async re-invocation: the moment you end your turn, the process exits and ANY background work you started is KILLED. Therefore:

\begin{itemize}
\item Run every computation SYNCHRONOUSLY in the FOREGROUND and BLOCK until it finishes (e.g. run \texttt{python score.py} directly and wait for it to return --- do NOT use \texttt{nohup ... \&}, \texttt{\&} backgrounding, \texttt{disown}, or detached processes).
\item Do NOT use ScheduleWakeup, completion-waiters, sleep-then-yield, or any "I'll resume when notified" pattern --- you will NOT be resumed.
\item Long foreground commands are fine within the time budget. Only end your turn AFTER \texttt{outputs/group\_assignment.csv} is written and verified on disk.
\end{itemize}
\end{selectionprompt}

\subsection{Multi-turn Tool-use Selection}
\label{app:prompt-bfcl}
\begin{selectionprompt}
Read \texttt{config.yaml} in the current directory first. It specifies: \texttt{base\_model} (HF id / path), \texttt{dataset\_path} (the full SFT candidate dataset, parquet), \texttt{n\_groups} (=5), \texttt{group\_size} (=50), and \texttt{max\_len}. Use exactly these --- do not hardcode other values.

We have that base model and the full SFT candidate dataset. A FIXED training recipe (documented in \texttt{recipe/README.md}, \texttt{recipe/sft.yaml} and \texttt{recipe/expand.py}; read them, do not modify or run them) will later fine-tune the base model separately on each of \texttt{n\_groups} data groups, with identical hyper-parameters; the only variable across runs is which data subset goes in.

\promptheading{Downstream benchmark (what the trained models will be scored on)}
Each trained model will act as a TOOL-USE AGENT in a stateful environment: given a set of function definitions and a user request, it must decide over several turns which functions to call and with what arguments, read the returned results, and continue until the request is fulfilled --- including cases where a needed function is missing, a required argument was never given, or the context is very long. It is scored by whether the environment ends in the correct state. You never see the benchmark tasks; you only know the evaluation is of this kind.

\promptheading{Your goal}
This is a SELECTION-AND-RANKING task. Predict, for every trajectory in the dataset, its expected value for SFT (how much it would improve the downstream benchmark if trained on), and use that prediction to SELECT \texttt{n\_groups} groups of exactly \texttt{group\_size} rows each, ordered so that: after training each group with the fixed recipe, benchmark score goes from high to low as the group index goes from 1 to n\_groups (group 1 best, group n\_groups worst).

You are FREE to design the valuation however you think works best. You may use a SINGLE signal or COMBINE MULTIPLE complementary signals (e.g. loss/perplexity, trajectory quality, task difficulty, tool-call correctness, efficiency, diversity/coverage of environments and tools, length effects, ...) into a composite score --- your choice. Don't feel obliged to use just one textbook metric; use whatever you believe will best achieve the objective below.

\promptheading{Objective (what you are optimizing)}
\begin{enumerate}
\item \textbf{PRIMARY --- make group 1 as strong as possible.} Group 1 should be the single most valuable \texttt{group\_size}-row subset you can assemble from the whole pool: after training on it alone, its benchmark score should be as HIGH as possible --- ideally beating a random subset of the same size. Put real effort into getting the very top group right; that is the headline result.
\item \textbf{ORDERING --- rank the groups.} Score should decrease from group 1 to group n\_groups (Spearman between group index and post-trained score -\textgreater{} -1). Place the most valuable trajectories in the lowest-numbered groups and the least valuable of your selection in the highest.
\end{enumerate}
\promptheading{Deliverable (the ONLY thing you produce)}
Write \texttt{outputs/group\_assignment.csv} with columns: \texttt{index,meta\_index,group\_id}

\begin{itemize}
\item \texttt{index}      = the 0-based ROW POSITION of the trajectory in the dataset file (config.dataset\_path): 0 for the first row, n\_rows-1 for the last.
\item \texttt{meta\_index} = the value of the \texttt{meta\_index} COLUMN of that same row. It is an identifier, NOT the row position, and differs from \texttt{index} for most rows. Write both: they are checked against each other and the file is rejected if any pair disagrees, so a mix-up surfaces immediately instead of silently selecting other trajectories.
\item \texttt{group\_id}   = an integer in 1..n\_groups
\item The file contains ONLY the selected rows: exactly n\_groups x group\_size lines (plus header), each selected index exactly once, exactly \texttt{group\_size} rows per group\_id. Unselected rows must NOT appear in the file.
\end{itemize}
Also write \texttt{outputs/method.md} briefly explaining your method and why. Do NOT produce a trained model. Do NOT run the training recipe.

\promptheading{\texttt{index} and \texttt{meta\_index} are DIFFERENT numbers --- read this before writing the file}
\texttt{index} is WHERE a trajectory sits in the file. \texttt{meta\_index} is an ID stored INSIDE the row, carried over from the corpus this pool was filtered out of. Because some trajectories were removed when the pool was built, the two drift apart: they agree for the first rows and then differ for about 96\% of the file, by up to 60. Concretely, the row at position 379 carries \texttt{meta\_index} 380.

In pandas terms: \texttt{index} is the position you would pass to \texttt{.iloc[]}; \texttt{meta\_index} is the column \texttt{df["meta\_index"]}. Reading the column when you need the position (or the reverse) selects DIFFERENT trajectories than the ones you scored, and most of those wrong values are themselves in range, so nothing would look broken.

For each selected trajectory write the pair exactly as the dataset has it: \texttt{index} = its row position, \texttt{meta\_index} = \texttt{df["meta\_index"].iloc[index]}.

\promptheading{Hard constraints}
\begin{enumerate}
\item INFERENCE-ONLY. You may only run forward passes of the base model (hidden states / logits are accessible). NO gradients, NO optimizer steps, NO training of any model (not even a small proxy). Gradient/optimizer calls are disabled in this sandbox and will raise.
\item The base model (config.base\_model) is READ-ONLY. Use it as-is; do not fine-tune or overwrite it.
\item Do NOT access any benchmark/test set. Only use the dataset at \texttt{config.dataset\_path}.
\item The n\_groups groups must be mutually disjoint and each contain exactly \texttt{group\_size} rows. They need NOT cover the dataset: rows outside your selection are left unused.
\item Work only in the current directory. Internet is available for packages if needed (transformers/datasets/torch/pyarrow are pre-installed).
\end{enumerate}
\promptheading{Data format}
Each row of the dataset is ONE multi-turn tool-use trajectory. Columns: \texttt{meta\_index} (an identifier carried over from the source corpus -- NOT the row position), \texttt{env\_id} (the simulated environment the trajectory was collected in), \texttt{task\_id}, \texttt{traj\_type} ("conversation" or "non\_conversation"), \texttt{steps} (number of assistant turns) and \texttt{messages}, a list of role/content dicts:

\begin{itemize}
\item \texttt{messages[0]} is the system message; it contains the environment instructions AND the function definitions available in that trajectory (Qwen3 \texttt{\textless{}tools\textgreater{}} format).
\item Then user and assistant messages alternate. A user message is either the user's request or one or more \texttt{\textless{}tool\_response\textgreater{}...\textless{}/tool\_response\textgreater{}} blocks returning what the previously called functions produced.
\item EVERY assistant message starts with its reasoning in \texttt{\textless{}think\textgreater{}...\textless{}/think\textgreater{}}, followed by text and/or \texttt{\textless{}tool\_call\textgreater{}\{"name": ..., "arguments": \{...\}\}\textless{}/tool\_call\textgreater{}} blocks.
\end{itemize}
The recipe does NOT train on a row as one sequence: it expands a row with \texttt{steps} assistant turns into \texttt{steps} samples, each supervising one assistant turn (with its \texttt{\textless{}think\textgreater{}}) given the preceding turns with their \texttt{\textless{}think\textgreater{}} blocks removed --- see \texttt{recipe/expand.py}, and use it (or match it exactly) when you compute anything per sample with the model chat template (tokenizer.apply\_chat\_template), truncating at \texttt{max\_len}.

\promptheading{Environment}
\begin{itemize}
\item A single H100 GPU (80 GB) is available. The dataset (config.dataset\_path), the base model (config.base\_model), and \texttt{recipe/} (read-only reference) are in the current directory.
\item Scale: about 9k trajectories averaging \textasciitilde{}18 turns and \textasciitilde{}6.5k tokens each when rendered; a full forward pass over the whole pool is feasible on this GPU but not free --- plan it.
\item TIME BUDGET: complete the task within 4 hours. Plan your computation to fit --- the deliverable must be written and verified before the budget runs out.
\item You operate fully autonomously; never ask for feedback. Just produce \texttt{outputs/group\_assignment.csv}.
\end{itemize}
\promptheading{Execution model (IMPORTANT --- single headless turn)}
You run as ONE non-interactive, single-shot invocation. There is NO async re-invocation: the moment you end your turn, the process exits and ANY background work you started is KILLED. Therefore:

\begin{itemize}
\item Run every computation SYNCHRONOUSLY in the FOREGROUND and BLOCK until it finishes (e.g. run \texttt{python score.py} directly and wait for it to return --- do NOT use \texttt{nohup ... \&}, \texttt{\&} backgrounding, \texttt{disown}, or detached processes).
\item Do NOT use ScheduleWakeup, completion-waiters, sleep-then-yield, or any "I'll resume when notified" pattern --- you will NOT be resumed.
\item Long foreground commands are fine within the time budget. Only end your turn AFTER \texttt{outputs/group\_assignment.csv} is written and verified on disk.
\end{itemize}
\end{selectionprompt}

\end{document}